\documentclass{article}

\PassOptionsToPackage{numbers}{natbib}

\usepackage[preprint]{neurips_2026}

\usepackage[utf8]{inputenc}
\usepackage[T1]{fontenc}

\usepackage{bm}
\usepackage{microtype}
\usepackage{graphicx}
\usepackage{caption}
\usepackage{subcaption}
\usepackage{placeins}
\usepackage{wrapfig}

\usepackage{amsmath}
\usepackage{amssymb}
\usepackage{amsfonts}
\usepackage{mathtools}
\usepackage{amsthm}

\usepackage{booktabs}
\usepackage{multirow}
\usepackage{tabularx}
\usepackage{array}
\usepackage{colortbl}
\usepackage{nicefrac}

\usepackage{enumitem}
\usepackage[ruled,vlined]{algorithm2e}

\usepackage[table,dvipsnames]{xcolor}
\usepackage{tcolorbox}
\tcbuselibrary{theorems}

\usepackage{url}

\definecolor{academicBlue}{rgb}{0.28,0.39,0.55}
\definecolor{academicBack}{rgb}{0.97,0.98,0.99}
\definecolor{opencolor}{RGB}{25,118,210}

\definecolor{mydarkred}{rgb}{0.6,0,0}
\definecolor{myblue}{HTML}{268BD2}
\definecolor{mygreen}{HTML}{658354}
\definecolor{orangeinplot}{HTML}{e29c7a}
\definecolor{purpleinplot}{HTML}{7676a4}
\definecolor{greeninplot}{HTML}{288308}

\usepackage{graphicx}
\usepackage{fontawesome5}

\usepackage[
colorlinks,
linkcolor=mydarkred,
urlcolor=RoyalBlue,
citecolor=myblue
]{hyperref}

\title{
Block-R1: Rethinking the Role of Block Size in Multi-domain Reinforcement Learning for \\
Diffusion Large Language Models
}

\author{
\begin{tabular}{c}
Yan Jiang, Ruihong Qiu, Zi Huang \\[0.6em]
The University of Queensland \\[0.3em]
\texttt{yan.jiang@uq.edu.au} \\[0.3em]
\faGithub\quad \textbf{Code:} 
\href{https://github.com/YanJiangJerry/Block-R1}
{\texttt{github.com/YanJiangJerry/Block-R1}} \\[0.3em]
\raisebox{-0.2em}{\includegraphics[height=1em]{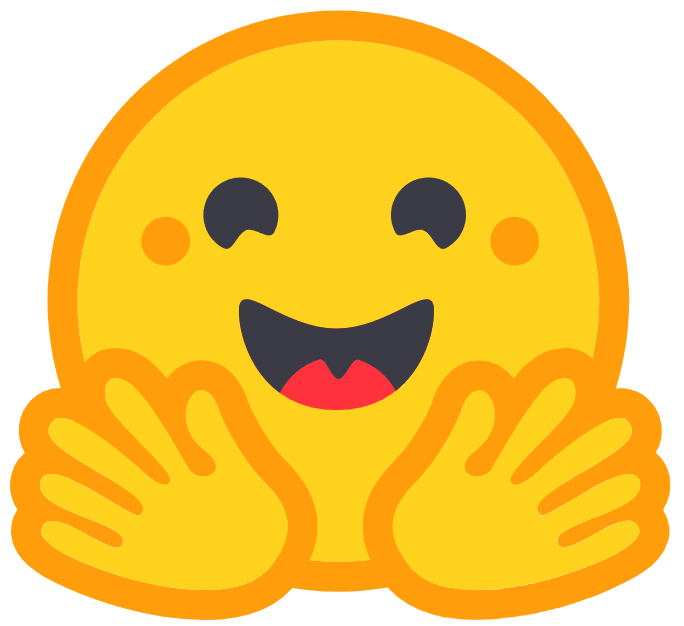}}
\quad \textbf{Dataset:}
\href{https://huggingface.co/datasets/YanJiangJerry/Block-R1-41K}
{\texttt{huggingface.co/datasets/YanJiangJerry/Block-R1-41K}}
\end{tabular}
}
\begin{document}
\maketitle

\begin{abstract}

Recently, reinforcement learning (RL) has been widely applied during post-training for diffusion large language models (dLLMs) to enhance reasoning with block-wise semi-autoregressive generation. Block size has therefore become a vital factor in dLLMs, since it determines the parallel decoding granularity and affects the rollout trajectories during RL optimisation, e.g., GRPO. Instead of investigating the effect of block size during inference on individual domains, this paper studies block size from a domain conflict perspective for dLLM RL post-training in multi-domain scenarios. The main contributions are: (1) a formulation of domain block size conflict in multi-domain RL for dLLMs, which will largely affect the post-training effectiveness for rollout-based RL methods; (2) a novel dataset, \textbf{Block-R1-41K} is constructed with a best-improved training block size for each sample, which also induces a Block Size Conflict Score to quantitatively measure the domain conflict; (3) a new benchmark, \textbf{Block-R1}, for flexible RL post-training for dLLMs in both single and cross domain; and (4) a simple yet powerful cross-domain post-training method with sample-level best-improved training block sizes. Extensive experiments on 13 distinct datasets, 7 latest RL algorithms, and various different dLLM backbones are covered in Block-R1. 
\end{abstract}

\section{Introduction}
Diffusion Large Language Models (dLLMs)~\cite{nie2025large, dream2025, 2025blockdiffusion, bie2025llada20, bie2026llada21, zhu2025llada} have recently emerged as a promising alternative to conventional autoregressive large language models by enabling parallel token generation through a block-based semi-autoregressive decoding mechanism~\cite{2025blockdiffusion}. Specifically, the sequence is divided into equal-length structured blocks, enabling parallel decoding of multiple tokens within each block while maintaining sequential generation across blocks. To further enhance block-based dLLMs, recent studies~\cite{zhao2025d1, tang2025wd1, rojas2025gdpo, zhong2026stabilizing, wang2026spg, ou2025espo} have attempted to employ reinforcement learning (RL), particularly Group Relative Policy Optimisation (GRPO), to post-train dLLMs with fixed block size configurations, yielding remarkable improvements especially on reasoning tasks~\cite{xiong2025minimalistapproachllmreasoning, tang2026multiplexthinking}.

\begin{figure*}[!t]
    \vspace{-0.4cm}
    \centering
    \includegraphics[width=0.9\linewidth]{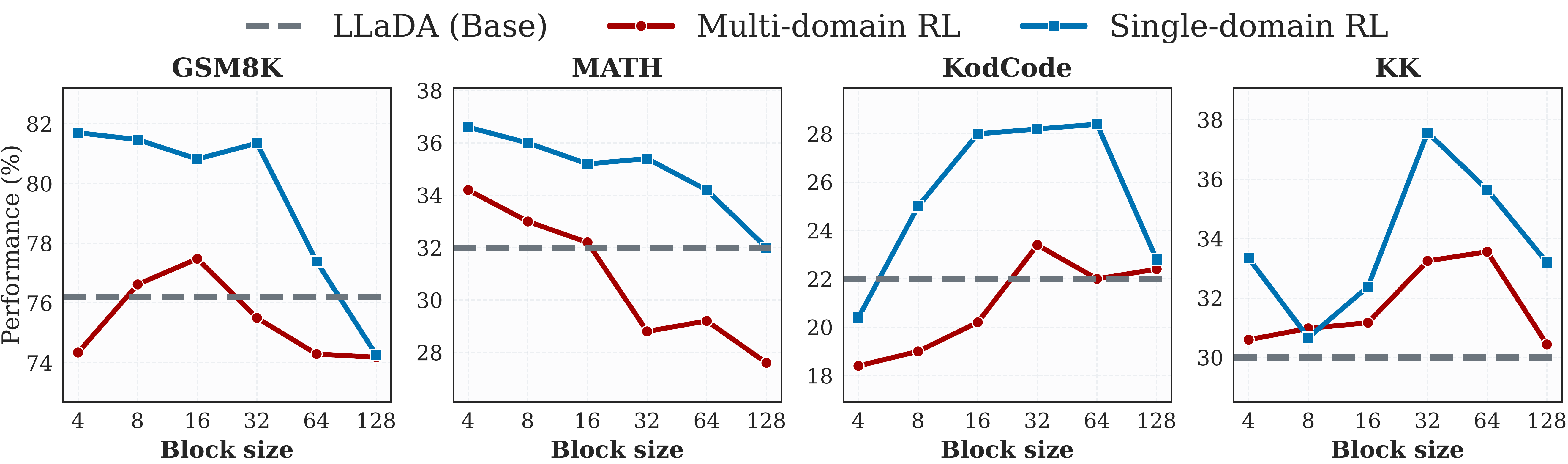}
    \vspace{-0.2cm}
    \caption{Motivation for Block-R1. Multi-domain RL refers to using all six domains as training domains during post-training. The recent competitive method StableDRL~\cite{zhong2026stabilizing} is used as the default base RL algorithm owing to its strong performance, while LLaDA-8B~\cite{nie2025large} is adopted as the dLLM backbone following most existing dLLM post-training methods~\cite{zhao2025d1, tang2025wd1, rojas2025gdpo}. The insights are twofold: \textbf{(1) RL post-training dLLMs on different domains prefer different block sizes. (2) Multi-domain post-training on all domains with the same block size (red line) results in degraded dLLM test performance compared with single-domain RL (blue line)}, and can even perform worse than the base model without RL (grey dashed line). Evidence by more dLLM-based RL methods is in Figure~\ref{fig:block_d1}}
    \vspace{-0.7cm}
    \label{fig:block_stable}
\end{figure*}

Despite their progress, existing dLLM-targeted RL methods remain largely limited to single-domain scenarios~\cite{DARE, zhao2025d1, tang2025wd1, rojas2025gdpo, wang2026spg, ou2025espo, zhong2026stabilizing}, where a dLLM is post-trained on one domain and evaluated within the same domain. This paradigm risks overfitting dLLMs to domain-specific patterns, thereby weakening their generalisation to unseen domains. To enhance dLLM generalisability across various domains, a straightforward approach is to conduct multi-domain RL post-training~\cite{cheng2025revisiting} using the same block size as usual. Nevertheless, our empirical observations in Figure~\ref{fig:block_stable} and theoretical insights in Section~\ref{sec:block_r1} reveal that severe block-size-related domain conflict exists, which hinders the effectiveness of dLLM multi-domain post-training. As shown in Figure~\ref{fig:block_stable}, RL post-training a dLLM on different domains prefers different block size configurations, while joint multi-domain post-training with the same block size results in degraded performance compared with single-domain RL, and can even perform worse than the base model. The underlying cause lies in the heterogeneity of dLLM parallel generation structures across domains, since distinct tasks inherently reside on different best-improved parallel decoding granularities. For example, mathematical reasoning tasks such as GSM8K may benefit from smaller blocks for fine-grained intermediate verification~\cite{chen2025bridging, cobbe2021training}, whereas puzzle-solving tasks, such as Sudoku, may favour larger blocks to preserve global format consistency across rows, columns, and subgrids~\cite{arel_sudoku}. When a one-for-all block size is imposed during multi-domain RL, all domains generate rollouts under the same block-based decoding structure for policy optimisation, even though their best-improved decoding granularities may differ. As a result, the policy update may be well-suited to some domains but sub-optimal for others. Therefore, block size is an important factor in multi-domain post-training of dLLMs rather than a fixed, unchanged decoding hyperparameter.

In light of the above discussion, in this paper, we propose \textbf{Block-R1}, a new benchmark for cross-domain RL post-training for dLLMs. The central objective of Block-R1 is to explicitly study and mitigate domain block size conflict by breaking the rigid unified block size constraint. Instead of treating block size as a fixed decoding hyperparameter, Block-R1 estimates sample-level best-improved training block sizes via empirical reward evaluation across various block sizes. Specifically, for each training sample, reward performance under different block sizes is assessed through a carefully designed teacher-student evaluation pipeline, and the derived best-improved training block size is embedded into the collected multi-domain dataset. This dataset design enables a simple yet powerful cross-domain training method for dLLMs, where rollout generation and policy updates are conditioned on sample-level block-size annotations. 13 distinct benchmark datasets, 7 latest RL algorithms, and various dLLM backbones are covered in Block-R1. Our main contributions are:
\begin{itemize}[leftmargin=*, itemsep=0pt, topsep=2pt, parsep=0pt, partopsep=0pt] 
\item \textbf{Insights into Domain Block Size Conflict:} We formally formulate domain block size conflict in multi-domain RL for dLLMs and provide theoretical insights showing that a fixed, one-for-all block size configuration inherently restricts global optimisation capacity across heterogeneous domains.
\item \textbf{Block-R1-41K Dataset:} We construct a novel block-based multi-domain dataset for dLLMs that assigns each sample a best-improved block size through a teacher-student evaluation pipeline.
\item \textbf{Cross-domain RL Benchmark for dLLMs:} We introduce Block-R1 as a comprehensive benchmark for cross-domain RL post-training for dLLMs, covering 13 distinct datasets, 7 latest RL algorithms, and various dLLM backbones.
\item \textbf{Sample-level Block-conditioned Training:} We provide a simple yet powerful cross-domain training method that uses sample-level best-improved training block sizes to enable block-conditioned policy updates across domains for dLLMs.
\end{itemize}

\section{Preliminaries}
\label{sec:preliminaries}
This section introduces the notation and objectives used throughout the paper. Let $\mathbf{x} = [x_1, x_2, \dots, x_L]$ denote a token sequence of length $L$. For block-based generation in dLLMs, the sequence is partitioned into $K$ non-overlapping blocks $\mathcal{B} = \{\mathbf{b}_1, \mathbf{b}_2, \dots, \mathbf{b}_K\}$. The index $k$ identifies the $k$-th block, and $|\mathbf{b}_k|$ denotes the number of token positions in that block. In conventional fixed-size block generation, a constant block size $c$ enforces $|\mathbf{b}_k| = c$ for all full blocks. Let $B = \{c_1, c_2, \dots, c_S\}$ denote a candidate set of block sizes.

\subsection{Diffusion Large Language Models}
\label{sec:dllm_training}

\paragraph{Training.} The training process of dLLMs operates by reversing a corruption process to reconstruct the original text sequence. The generative model is parametrised by $\theta$, and the original clean sequence is denoted by $\mathbf{x}_0=(x_0^1,\dots,x_0^L)$. At diffusion time $t \in [0,1]$, the corrupted sequence is denoted by $\mathbf{x}_t$, where $t=1$ represents the fully masked sequence. The denoising cross-entropy loss is defined as:
\begin{align}
\label{eq:diff_loss}
\mathcal{L}(\theta) = - \mathbb{E}_{t \sim \mathcal{U}[0,1], \mathbf{x}_t} \Bigg[ \frac{1}{t} \sum_{i=1}^{L} \mathbb{I}(x_t^i = \texttt{mask}) \log \pi_\theta(x_0^i \mid \mathbf{x}_t) \Bigg].
\end{align}
In this formulation, $t \sim \mathcal{U}[0,1]$ represents continuous diffusion time, and $\mathbf{x}_t$ denotes the corrupted sequence at time $t$. The term $\frac{1}{t}$ weights the masked-token reconstruction loss across diffusion times. The indicator function $\mathbb{I}(\cdot)$ confines the loss computation to masked tokens, while $\pi_\theta(x_0^i \mid \mathbf{x}_t)$ denotes the probability predicted by the dLLM for the original token $x_0^i$ conditioned on the corrupted context.

\paragraph{Block-based Parallel Generation Structure.}
\label{sec:dllm_inference}
Most existing dLLMs adopt a semi-autoregressive generation by partitioning the sequences into discrete blocks of equal length. Unlike autoregressive models that generate tokens sequentially one by one, dLLMs generate blocks sequentially while denoising tokens within each block in parallel. Such a block-based generation structure is determined by the block size $c$, since $c$ determines how many tokens are jointly denoised within each block and therefore controls the granularity of the dLLM block-based parallel generation structure.

For fixed-size block generation, each block $\mathbf{b}_k$ is determined by a constant block size $c$:
\begin{equation} 
\label{eq:block_indices}
    \mathbf{b}_k = \{i \mid (k-1)c + 1 \leq i \leq \min(kc, L)\}.
\end{equation}
For each block $\mathbf{b}_k$, the dLLM performs diffusion denoising steps to recover all masked tokens $x_i \in \mathbf{b}_k$ in parallel. A larger value of $c$ allows more tokens to be denoised simultaneously, leading to a coarser generation structure. In contrast, a smaller value of $c$ restricts each parallel generation unit to fewer tokens, leading to a finer generation structure close to auto-regressive paradigm. Therefore, the block size $c$ determines the structural granularity of dLLM generation.


\subsection{Reinforcement Learning for dLLMs}

Reinforcement Learning functions as a powerful post-training mechanism to enhance dLLM reasoning capabilities by optimising trajectory-level rewards~\cite{yang2025mmada, wang2025traceRL}. Recent studies apply policy optimisation algorithms, particularly Group Relative Policy Optimisation (GRPO)~\cite{shao2024deepseekmath, zhao2025d1, tang2025wd1, rojas2025gdpo, he2025mdpo, zhong2026stabilizing, wang2026spg, ou2025espo}, to dLLM domains. Generally, the diffusion-based GRPO objective~\cite{zhao2025d1} can be formulated as:
\begingroup
\small
\begin{equation} 
\label{eq:grpo_obj}
\begin{aligned}
\mathcal{J}_{\mathrm{GRPO}}(\theta,c)
=
\mathbb{E}
\Bigg[
\frac{1}{GL}
\sum_{g=1}^{G}
\sum_{i=1}^{L}
\min\Big(
r_g^i(\theta)\hat{A}_g,
\mathrm{clip}(r_g^i(\theta),1-\epsilon,1+\epsilon)\hat{A}_g
\Big)
-
\beta D_{\mathrm{KL}}(\pi_\theta^{(c)} \| \pi_{\mathrm{ref}}^{(c)})
\Bigg],
\end{aligned}
\end{equation}
\endgroup
where the expectation is over prompts $x \sim \mathcal{D}$. Let $\tau_{1:G}=\{\tau_g\}_{g=1}^{G}$ denote $G$ trajectories sampled from the old block-conditioned policy $\pi_{\mathrm{old}}^{(c)}(\cdot \mid x)$. Here, $\tau_g$ denotes the $g$-th block-based trajectory generated with block size $c$, and $o_g^i$ denotes its $i$-th generated token. The importance sampling ratio is defined as $r_g^i(\theta)=\pi_\theta(o_g^i \mid \mathbf{x}_{g,t})/\pi_{\mathrm{old}}(o_g^i \mid \mathbf{x}_{g,t})$, where $\mathbf{x}_{g,t}$ is the intermediate denoising context. The advantage $\hat{A}_g$ is computed from the trajectory rewards $\{R(\tau_j)\}_{j=1}^{G}$ within the same group. The parameter $\epsilon$ denotes the clipping bound, and $\beta$ controls the KL regularisation term.

\paragraph{\textbf{\textit{Remark 1: Block-Dependent Policy Optimisation for dLLMs}}}
\label{rem:block_rl}
The conditioning of the RL objective for dLLMs on block size highlights a structural dependency. The objective $\mathcal{J}_{\text{GRPO}}(\theta, c)$ is explicitly conditioned on $c$ because the trajectories used to compute rewards and group advantages $\hat{A}_g$ are sampled from the block-conditioned policy $\pi_{\mathrm{old}}^{(c)}(\cdot \mid x)$ and governed by block-wise decoding. Consequently, changing $c$ alters the rollout trajectory distribution, the sequence evolution path, the resulting reward distribution, and the group-relative advantage estimates. Such block-dependent optimisation behaviour forms the basis of the training block size conflict studied in this paper.

\section{Block Size Conflict across Domains in RL Post-training for dLLMs}
\label{sec:block_r1}
Multi-domain RL for dLLMs is challenging mainly due to the existence of domain conflict, where block size mismatch is a key factor. Different reasoning domains may favour different decoding block sizes for effective parallel generation during RL optimisation. We formulate this conflict as follows:
\begin{tcolorbox}[
  colframe=academicBlue, 
  colback=academicBack,   
  coltitle=white,
  fonttitle=\bfseries, 
  title=Definition 1: Domain Block Size Conflict Challenge, 
  boxsep=1pt, left=1pt, right=2pt, top=1pt, bottom=1pt,
  arc=1mm, auto outer arc,
  label=def:block_conflict
]
Let $\mathcal{D}_i$ and $\mathcal{D}_j$ represent two distinct reasoning domains, and let $J_k(\theta, c) = \mathbb{E}_{x \sim \mathcal{D}_k, \tau \sim \pi_\theta^{(c)}(\cdot \mid x)}[R(\tau)]$ denote the expected reward for domain $\mathcal{D}_k$ evaluated under the block-based generation trajectory by size $c \in B$. Domain block size conflict occurs when the reward-preferred dLLM decoding granularities for these domains are mutually exclusive:
\begin{equation} \label{eq:domain_conflict}
\begin{aligned}
\arg\max_{c \in B} J_i(\theta, c) \cap \arg\max_{c \in B} J_j(\theta, c) = \emptyset.
\end{aligned}
\end{equation}

Intuitively, this fundamental incompatibility indicates that a globally fixed block size cannot simultaneously maximise the reasoning objectives required by structurally diverse domains.
\end{tcolorbox}

To investigate how this domain-level conflict affects multi-domain RL optimisation, we introduce two assumptions that connect sample-level rewards with domain-level dLLM post-training objectives and characterise the divergence of block-size preferences across domains:
\paragraph{\textbf{\textit{Assumption 1: (Objective Alignment)}}}
Let $A_\theta(x,c)=\mathbb{E}_{\tau \sim \pi_\theta^{(c)}(\cdot \mid x)}[R(\tau)]$ denote the expected reward of sample $x$ under block size $c$ and model parameter $\theta$. The sample-level reward is assumed to be aligned with the domain-level objective in Definition~\ref{def:block_conflict}, i.e., $J_k(\theta,c)=\mathbb{E}_{x \sim \mathcal{D}_k}[A_\theta(x,c)]$.
\paragraph{\textbf{\textit{Assumption 2: (Preference Divergence)}}}
For a dLLM with parameter $\theta$, domain block size conflict induces preference divergence across domains, i.e., there exist two domains $\mathcal{D}_i$ and $\mathcal{D}_j$ whose reward-preferred block sizes are disjoint: $\arg\max_{c \in B} J_i(\theta,c) \cap \arg\max_{c \in B} J_j(\theta,c)=\emptyset$.

Under these assumptions, fixed-size block generation leads to the multi-domain sub-optimality as:
\begin{tcolorbox}[
  colframe=academicBlue, 
  colback=academicBack,   
  coltitle=white,
  fonttitle=\bfseries, 
  title=Theorem 1: Sub-optimality of Multi-domain RL for dLLMs with Fixed-size Block, 
  boxsep=1pt, left=1pt, right=1pt, top=1pt, bottom=1pt,
  arc=1mm, auto outer arc,
  label=thm:conflict
]
Let $c_0 \in B$ denote any globally fixed block size uniformly applied across all domains $\mathcal{D}_k$, and let $\lambda_k>0$ represent normalised sampling weights satisfying $\sum_{k=1}^{M}\lambda_k=1$. Assume $B$ is non-empty and finite so that all maxima over $B$ exist. For any fixed dLLM $\theta$ satisfying Assumptions~1 and 2, the aggregated multi-domain RL objective under the fixed block size $c_0$ is strictly bounded as:
\begin{equation}
\label{eq:theorem_inequality}
\begin{aligned}
\sum_{k=1}^{M} \lambda_k J_k(\theta, c_0) 
< 
\sum_{k=1}^{M} \lambda_k \max_{c \in B} J_k(\theta, c).
\end{aligned}
\end{equation}
\end{tcolorbox}
The proof is deferred to Appendix~\ref{sec:theoretical_proof}. Intuitively, Theorem~\ref{thm:conflict} shows that a one-for-all block size is structurally sub-optimal for dLLM multi-domain RL when domains exhibit preference divergence. A natural solution is to assign one block size to each domain, yet this remains coarse because different sample tasks within the same domain can still reside on different parallel decoding granularities. Therefore, Block-R1 adopts sample-level block-size allocation as in Section~\ref{sec:dataset}, allowing each training instance to generate trajectories under its most suitable block structure.

\section{Block-R1-41K: Block-based Multi-domain Dataset for dLLMs}
\label{sec:dataset}
This section introduces the Block-R1-41K dataset to break the rigid block size constraint from existing RL methods for effective multi-domain post-training in Block-R1. The detailed algorithm and complexity analysis are in Algorithm~\ref{app:algorithm} and Appendix~\ref{app:complexity}, with a detailed pipeline in Figure~\ref{fig:main}.

\paragraph{\textbf{\textit{Stage 1: Cross-Domain Source Selection}}}
The initial phase aims to select high-quality reasoning benchmark datasets from various domains to establish a clean foundation for multi-domain dLLM post-training. To ensure data quality, the source selection incorporates popular reasoning benchmarks following \textit{d1}~\cite{zhao2025d1}. The post-training and evaluation protocols in \textit{d1} remain confined to the mathematical domain. To ensure comprehensive evaluation, the source collection is extended to five general domains as shown in Table \ref{tab:dataset_stats}. Benchmarks with official training sets are selected as data sources, while datasets without suitable training splits are retained for evaluation only.

\paragraph{\textbf{\textit{Stage 2: Cross-Domain Reward Design}}}
Different domains require distinct reward functions due to their heterogeneous output formats and task objectives. To support unified multi-domain RL post-training, Block-R1 constructs a shared reward design principle across all reasoning tasks. This formulation consists of three complementary reward components:

\begin{itemize}[leftmargin=*, itemsep=0pt, topsep=2pt, parsep=0pt, partopsep=0pt]
\item \textbf{Format-based rewards} enforce a canonical output structure for stable automated parsing. The model receives credit when its generation is enclosed within predefined machine-extractable tags such as \texttt{<reasoning>} and \texttt{<answer>}.
\item \textbf{Accuracy-based rewards} evaluate task success through domain-specific verification after answer extraction. Mathematical reasoning uses equivalence-aware symbolic comparison, code generation uses execution-based unit tests, and logical puzzles use fractional correctness over assignments.
\item \textbf{Constraint-based rewards} provide continuous supervision for well-formed but incomplete outputs to reduce sparse feedback. They grant partial credit for syntactic validity, partial correctness, or valid structural mappings, improving reward density and learning stability.
\end{itemize}
This decomposed reward design establishes comparable learning signals across diverse domains while preserving the natural evaluation criteria of each reasoning task.

\paragraph{\textbf{\textit{Stage 3: Filtering by Teacher and Student Models}}}

A primary challenge for dLLM post-training is selecting samples that provide useful learning signals. To address this, Block-R1 adopts a teacher-student block-based comparison. The teacher model represents a stronger reachable reasoning state after post-training, while the student model represents the base capability before RL. Samples that exhibit a meaningful teacher-student improvement gap are considered valuable because they indicate reachable learning potential.

Specifically, difficulty filtering is conducted where a strong model, inclusionAI/LLaDA2.0-mini (16B), serves as the teacher, and a weak model, GSAI-ML/LLaDA-8B-Base, serves as the student. Both models are evaluated over the full candidate block-size set $B$. Let $A_T(x,c)$ and $A_S(x,c)$ denote the teacher and student expected rewards under block size $c$, respectively. Block-R1 removes:
\begin{itemize}[leftmargin=*, itemsep=0pt, topsep=2pt, parsep=0pt, partopsep=0pt]
\item \textbf{Overly easy samples}, where the student already obtains full reward under all candidate block sizes.
\item \textbf{Overly hard samples}, where the teacher obtains zero reward under all candidate block sizes.
\item \textbf{Anomaly samples}, where the teacher cannot outperform the student in the maximum reward score across all candidate block sizes, possibly due to parsing errors or reward evaluation noise.
\end{itemize}

This filtering process ensures that subsequent RL focuses on samples with non-trivial and reachable improvement potential, rather than wasting computation on trivial or unsolvable instances.

\paragraph{\textbf{\textit{Stage 4: Best-improved Training Block Size Selection}}}
Motivated by the insight in Theorem~\ref{thm:conflict}, Block-R1 extends block-size adaptation from the domain level to the sample task level. Rather than selecting one block size per domain, Block-R1 assigns the best-improved block size to each individual sample. To achieve that, this stage introduces the sample-Level best-improved block Size selection:
\begin{tcolorbox}[
  colframe=academicBlue, 
  colback=academicBack,   
  coltitle=white,
  fonttitle=\bfseries, 
  title=Definition 3: Sample-Level Best-Improved Block Size Selection, 
  boxsep=1pt, left=1pt, right=1pt, top=1pt, bottom=1pt,
  arc=1mm, auto outer arc,
  label=def:best-improved_block
]
Let $\theta_T$ denote the frozen teacher dLLM and $\theta_S$ denote the frozen student dLLM. The teacher and student expected rewards under block size $c$ can be defined as:
\begin{equation}
\label{eq:teacher_student_rewards}
\begin{aligned}
A_{\theta_T}(x,c)=\mathbb{E}_{\tau \sim \pi_{\theta_T}^{(c)}(\cdot \mid x)}
\left[
R(\tau)
\right],
\quad
A_{\theta_S}(x,c)=\mathbb{E}_{\tau \sim \pi_{\theta_S}^{(c)}(\cdot \mid x)}
\left[
R(\tau)
\right].
\end{aligned}
\end{equation}
Therefore, we define the sample-level best-improved training block size as:
\begin{equation}
\label{eq:improvement_and_best-improved_c}
\begin{aligned}
c_x^*
=
\arg\max_{c \in B}
\Delta(x,c),
\quad
\text{where }
\Delta(x,c)
=
A_{\theta_T}(x,c)-A_{\theta_S}(x,c).
\end{aligned}
\end{equation}
\end{tcolorbox}
Intuitively, Block-R1 selects the block size that maximises the teacher-student improvement gap, so that each training instance is paired with the decoding granularity where the student dLLM has the largest reachable improvement. In Appendix~\ref{app:exp}, we visualise the block size distribution and the reward improvement across domains, further supporting that block size is an important factor affecting reward improvement and that different domains prefer different training block sizes.

\paragraph{\textbf{\textit{Stage 5: Sample-level Block-conditioned Multi-domain dLLM Post-training}}}
The final stage uses the selected sample-level block annotations to enable multi-domain RL post-training with adaptive block-based generation. Let $\mathcal{X}_k$ denote the sample pool from the $k$-th source domain, where each sample $x \in \mathcal{X}_k$ has been assigned its best-improved block size $c_x^*$. To balance heterogeneous domain data, an equal number of training instances is sampled from each source domain as in Appendix~\ref{app:reproduce}. Let $\widetilde{\mathcal{X}}_k \subseteq \mathcal{X}_k$ denote the samples from the $k$-th source domain. Block-R1-41K is assembled as:
\begin{equation} 
\label{eq:dataset_r1}
\begin{aligned}
\mathcal{D}^{\mathrm{Block}\text{-}\mathrm{R1-41K}}
=
\{(x, c_x^*) \mid x \in \cup_{k=1}^{M} \widetilde{\mathcal{X}}_k \}.
\end{aligned}
\end{equation}

The final dataset contains more than 41K high-quality multi-domain training samples, with each sample paired with a best-improved training block size. During RL post-training, Block-R1 replaces the globally fixed block-conditioned policy $\pi_\theta^{(c)}$ in Equation~\ref{eq:grpo_obj} with the sample-specific block-conditioned policy $\pi_\theta^{(c_x^*)}$. Therefore, the model generates rollout trajectories as $\pi_\theta^{(c_x^*)}(\tau \mid x)$, allowing different training samples to use different decoding granularities. This design conditions policy optimisation on both the semantic input and the sample-specific decoding granularity, thereby mitigating domain block size conflict without enforcing a unified fixed block size across all domains.

\section{Measuring Domain Block Size Conflict in Multi-domain RL for dLLMs}
To measure training block size conflict between domains during dLLM multi-domain RL post-training, we establish a distributional metric based on the best-improved training block sizes collected in Stage 4. Unlike the strict structural condition in Theorem~\ref{thm:conflict}, this metric serves as an empirical diagnostic measurement. It measures the divergence between domain-level distributions of best-improved training block sizes by the teacher-student improvement criterion as:
\begin{tcolorbox}[
  colframe=academicBlue, 
  colback=academicBack,   
  coltitle=white,
  fonttitle=\bfseries, 
  title=Definition 4: Domain Block Size Preference Distribution, 
  boxsep=1pt, left=1pt, right=1pt, top=1pt, bottom=1pt,
  arc=1mm, auto outer arc,
  label=def:training_block_dist
]
Let $\mathcal{D}_k$ denote the $k$-th reasoning domain, and let $c_x^*$ denote the best-improved training block size of sample $x$ obtained from the teacher-student improvement criterion in Equation~\ref{eq:improvement_and_best-improved_c}. The domain-level training block size preference distribution can then be defined as:
\begin{equation} 
\label{eq:domain_dist}
\begin{aligned}
P_k^{\mathrm{train}}(c)
=
\mathbb{P}(c_x^* = c \mid x \sim \mathcal{D}_k).
\end{aligned}
\end{equation}
\end{tcolorbox}
Intuitively, $P_k^{\mathrm{train}}(c)$ represents the probability that a sample from domain $\mathcal{D}_k$ selects block size $c$ as its most suitable training block size for RL post-training. Based on it, the Block Size Conflict Score can be measured by the Wasserstein distance~\cite{wasserstein}. The rationale is that block sizes are ordered decoding granularities rather than unordered categories, and the Wasserstein distance captures both distributional mismatch and displacement along the block-size axis. The score is formally defined as:
\begin{tcolorbox}[
  colframe=academicBlue, 
  colback=academicBack,   
  coltitle=white,
  fonttitle=\bfseries, 
  title=Definition 5: Domain Block Size Conflict Score, 
  boxsep=1pt, left=1pt, right=1pt, top=1pt, bottom=1pt,
  arc=1mm, auto outer arc,
  label=def:bcs
]
The Domain Block Size Conflict Score (BCS) between two domains (i,j) is defined as the Wasserstein distance between their training block preference distributions:
\begin{equation} 
\label{eq:bcs}
\begin{aligned}
\mathrm{BCS}(\mathcal{D}_i, \mathcal{D}_j)
=W_1\left(P_i^{\mathrm{train}}, P_j^{\mathrm{train}}\right)
=\sum_{s=1}^{S-1}
\left|
F_i(c_s)-F_j(c_s)
\right|
(c_{s+1}-c_s),
\end{aligned}
\end{equation}
where $c_s$ denotes the $s$-th candidate block size, and $F_i(c_s)$ and $F_j(c_s)$ denote the cumulative distribution functions of $P_i^{\mathrm{train}}$ and $P_j^{\mathrm{train}}$, respectively.
\end{tcolorbox}

Intuitively, a larger BCS indicates that two domains prefer substantially different best-improved training block sizes under the teacher-student improvement criterion. This means that their samples tend to expose the largest reachable improvement under different parallel decoding granularities, making a one-for-all block size less suitable for multi-domain RL post-training. In Section~\ref{sec:bcs} and Appendix~\ref{app:bcs}, we visualise the training block size divergence across domains and show that stronger BCS is associated with weaker fixed-block multi-domain post-training performance.

\section{Benchmarking Cross-domain Reinforcement Learning for dLLMs}
\vspace{-0.4cm}
Existing RL methods for dLLMs usually develop upon their own training and evaluation protocols, making consistent comparison difficult across algorithms, domains, and decoding configurations. To systematically evaluate single-domain, cross-domain, and multi-domain RL post-training performance for dLLMs, we construct a comprehensive benchmark, \textbf{Block-R1}. Block-R1 provides a unified post-training and evaluation pipeline covering 7 recent RL for dLLMs algorithms and more than 13 benchmarks, as in Table~\ref{tab:dataset_stats}, enabling systematic evaluation of all methods under a consistent setting.
\textbf{Setup.} Block-R1 evaluates cross-domain RL post-training across 13 benchmarks, including code generation, mathematical reasoning, logical puzzles, general capabilities, and advanced reasoning tasks. It supports 10 dLLM backbones, including LLaDA-8B-Base, LLaDA-8B-Instruct, LLaDA-1.5, LLaDA2.0-mini, LLaDA2.1-mini, Dream-v0-Base-7B, Dream-v0-Instruct-7B, SDAR-8B-Chat-b32, TraDo-8B-Instruct, and TraDo-8B-Thinking. Unless otherwise specified, LLaDA-8B-Instruct~\cite{nie2025large} is used as the default backbone following most prior methods~\cite{zhao2025d1, tang2025wd1, rojas2025gdpo, wang2026spg, ou2025espo, zhong2026stabilizing}. Zero-shot pass@1 accuracy is reported under a consistent setting, with default generation length 256, diffusion step 128, and fixed-block baseline block size 32 following most prior methods~\cite{zhao2025d1, tang2025wd1, rojas2025gdpo}.

\textbf{Baseline.} Our baselines include diverse RL methods for dLLMs, dynamic block size inference methods, and supervised fine-tuning. For RL frameworks, Block-R1 supports both token-level methods, including \textit{Diffu}-GRPO~\cite{zhao2025d1}, \textit{d1}~\cite{zhao2025d1}, and \textit{wd1}~\cite{tang2025wd1}, and sequence-level methods, including GDPO~\cite{rojas2025gdpo}, MDPO~\cite{he2025mdpo}, ESPO~\cite{ou2025espo}, and StableDRL~\cite{zhong2026stabilizing}. Block-R1 also supports dynamic block size methods such as \textit{b1}~\cite{b1}. SFT follows the denoising cross-entropy objective in LLaDA~\cite{nie2025large}.

\textbf{Implementation.} Unless otherwise specified, experiments are conducted on four AMD MI300X GPUs, each with 192GB of memory. The best checkpoint accuracy is reported in zero-shot setting following prior methods~\cite{zhao2025d1, tang2025wd1, rojas2025gdpo}. StableDRL~\cite{zhong2026stabilizing} is adopted as the default RL algorithm for all experiments due to its robust empirical performance. More implementation details are in Appendix~\ref{app:reproduce}.

\begin{table*}[t]
\centering
\caption{\textbf{Evaluation of RL cross-domain generalisation for dLLMs.}
LLaDA-8B-Instruct~\cite{nie2025large} serves as the default dLLM following most prior methods~\cite{zhao2025d1, tang2025wd1, rojas2025gdpo, zhong2026stabilizing}. It is post-trained with different source domains (Rows) and tested against 13 target-domain benchmarks (Columns). StableDRL~\cite{zhong2026stabilizing} is the default RL algorithm.
\textbf{Cell Colors:} \colorbox{academicBlue!20}{Blue cells} denote base performance without RL. \colorbox{orange!20}{Orange cells} denote In-domain RL where training and testing domains are the same. \colorbox{purple!20}{Purple cells} denote Multi-domain RL on all training domains in Table~\ref{tab:dataset_stats}. White cells denote cross-domain performance. 
\textbf{Vanilla} directly performs multi-domain training on all domains with a fixed block size of 32. The best checkpoint accuracy are reported under a consistent zero-shot setting for all domains and is highlighted by {\color{purple}\textbf{best}} and {\color{teal}\underline{runner-up}}. \textbf{More detailed results for different dLLMs, other RL methods and single domain RL benchmarks are provided in Table \ref{tab:multi_backbone_benchmark}, \ref{table:single_benchmark} and \ref{table:block_r1_single_domain} in Appendix~\ref{app:exp}.}}
\label{tab:full_cross_domain_llada}
\vspace{-0.2cm}
\resizebox{1\linewidth}{!}{
\begin{tabular}{lccccccccccccc}
\toprule
\multirow{3}{*}{\textbf{Source / Target}} 
& \multicolumn{3}{c}{\textbf{Mathematical Reasoning}} & \multicolumn{3}{c}{\textbf{Code Generation}} & \multicolumn{2}{c}{\textbf{Puzzle Solving}} & \multicolumn{3}{c}{\textbf{General Capabilities}} & \multicolumn{2}{c}{\textbf{Advanced Reasoning}} \\
\cmidrule(lr){2-4} 
\cmidrule(lr){5-7} 
\cmidrule(lr){8-9} 
\cmidrule(lr){10-12}
\cmidrule(lr){13-14}
& \textbf{Countdown} & \textbf{GSM8K} & \textbf{MATH500} & \textbf{HumanEval} & \textbf{MBPP} & \textbf{KodCode} & \textbf{Sudoku} & \textbf{KK} & \textbf{HellaSwag} & \textbf{MMLU} & \textbf{ARC-E} & \textbf{MMLU-Pro} & \textbf{ARC-C} \\

\midrule
\rowcolor{academicBlue!20} \textbf{LLaDA (No RL)} & 16.80 & 76.19 & 32.00 & 28.66 & {\color{teal}\underline{33.60}} & 22.00 & 7.81 & 30.00 & 56.21 & 53.46 & 76.98 & 33.79 & 73.21 \\
\rowcolor{academicBlue!20} \quad \textbf{+ SFT} & 15.23 & 75.59 & 32.20 & 28.05 & 30.40 & 19.40 & 8.20 & 32.00 & 55.65 & 55.58 & 77.65 & 34.03 & 69.80 \\

\midrule
\textbf{Math Reasoning}\\
\quad \textbf{Countdown} & \cellcolor{orange!20}{\color{teal}\underline{\textbf{58.98}}} & 70.74 & 28.40 & 26.83 & 31.00 & 22.60 & 7.81 & 24.57 & 53.38 & 54.24 & 78.16 & 30.91 & 73.55 \\
\quad \textbf{GSM8K} & 16.02 & \cellcolor{orange!20}{\color{purple}\textbf{81.35}} & 32.60 & 27.44 & 32.20 & 22.60 & 10.55 & 30.57 & 58.07 & 56.22 & 80.51 & 33.96 & 75.51 \\
\quad \textbf{MATH500} & 14.06 & 77.33 & \cellcolor{orange!20}{\color{teal}\underline{\textbf{35.40}}} & 27.44 & 33.00 & 23.60 & 10.16 & 32.14 & 57.73 & 54.01 & 79.12 & 35.21 & {\color{teal}\underline{77.47}} \\

\midrule
\textbf{Code Generation}\\
\quad \textbf{KodCode} & 14.45 & 74.15 & 32.00 & \cellcolor{orange!20}{\color{teal}\underline{\textbf{31.10}}} & \cellcolor{orange!20}{\color{purple}\textbf{34.80}} & \cellcolor{orange!20}{\color{teal}\underline{\textbf{28.20}}} & 12.11 & 27.43 & 54.79 & 53.28 & {\color{teal}\underline{82.07}} & 33.84 & 75.94 \\

\midrule
\textbf{Puzzle Solving} \\
\quad \textbf{Sudoku} & 6.64 & 68.23 & 30.40 & 26.22 & 32.40 & 23.40 & \cellcolor{orange!20}{\color{teal}\underline{\textbf{25.39}}} & 33.29 & 57.24 & {\color{teal}\underline{56.77}} & 80.09 & {\color{teal}\underline{36.50}} & 73.46 \\
\quad \textbf{KK} & 8.98 & 70.28 & 29.60 & 24.39 & 29.80 & 20.80 & 15.63 & \cellcolor{orange!20}{\color{teal}\underline{\textbf{37.57}}} & {\color{teal}\underline{58.25}} & 55.62 & 81.10 & 34.89 & 74.32 \\

\midrule
\textbf{Multi-Domain RL}\\
\quad \textbf{Vanilla} & \cellcolor{purple!20}30.08 & \cellcolor{purple!20}57.24 & \cellcolor{purple!20}28.20 & \cellcolor{purple!20}24.39 & \cellcolor{purple!20}24.40 & \cellcolor{purple!20}22.60 & \cellcolor{purple!20}9.77 & \cellcolor{purple!20}30.14 & \cellcolor{purple!20}52.10 & \cellcolor{purple!20}52.06 & \cellcolor{purple!20}73.95 & \cellcolor{purple!20}29.79 & \cellcolor{purple!20}65.87 \\
\quad \textbf{Block-R1} & \cellcolor{purple!20}{\color{purple}\textbf{62.11}} & \cellcolor{purple!20}{\color{teal}\underline{80.74}} & \cellcolor{purple!20}{\color{purple}\textbf{35.80}} & \cellcolor{purple!20}{\color{purple}\textbf{34.76}} & \cellcolor{purple!20}{\color{purple}\textbf{34.80}} & \cellcolor{purple!20}{\color{purple}\textbf{28.60}} & \cellcolor{purple!20}{\color{purple}\textbf{26.95}} & \cellcolor{purple!20}{\color{purple}\textbf{50.14}} & \cellcolor{purple!20}{\color{purple}\textbf{64.07}} & \cellcolor{purple!20}{\color{purple}\textbf{62.22}} & \cellcolor{purple!20}{\color{purple}\textbf{90.53}} & \cellcolor{purple!20}{\color{purple}\textbf{37.96}} & \cellcolor{purple!20}{\color{purple}\textbf{82.51}} \\

\bottomrule
\end{tabular}
}
\vspace{-0.8cm}
\end{table*}
\subsection{Main Results}
\label{exp:main}
\vspace{-0.3cm}
Table~\ref{tab:full_cross_domain_llada} reports the cross-domain generalisation results of RL post-training for dLLMs. The insights are: \textbf{(1) Existing RL methods for dLLMs are effective in the single-domain setting but remain limited in cross-domain generalisation.} As shown by the orange cells, models post-trained on one source domain usually achieve strong performance on the same target domain. For example, the model post-trained on GSM8K achieves the best performance on GSM8K, but yields sub-optimal performance on many other domains, such as Countdown and HumanEval. Similarly, models post-trained on Countdown, KodCode, Sudoku, or KK improve their corresponding in-domain performance, but often degrade performance on other mathematical reasoning, code generation, and puzzle-solving benchmarks. This indicates that single-domain RL may overfit the model to domain-specific reasoning patterns and fails to provide stable generalisation across heterogeneous domains. \textbf{(2) Vanilla multi-domain RL with a unified fixed block size underperforms single-domain RL by a large margin.} Although multi-domain post-training exposes the model to samples from all training domains, the Vanilla baseline still performs poorly on most benchmarks. For example, Vanilla obtains only 30.08 on Countdown, 57.24 on GSM8K, and 24.39 on HumanEval, which are substantially lower than the corresponding in-domain RL results. This suggests that simply randomly mixing domains is insufficient for dLLM RL post-training as different domains inherently contain structural conflicts, especially block size conflict, and enforcing one fixed block size across all domains restricts the effectiveness of multi-domain rollout generation and policy optimisation on most domains. \textbf{(3) Block-R1 substantially improves multi-domain RL by assigning each training sample its own best-improved block size.} Compared with Vanilla multi-domain RL, Block-R1 achieves remarkable gains across all target domains, including mathematical reasoning, code generation, puzzle solving, general capabilities, and advanced reasoning. For example, Block-R1 improves Countdown from 30.08 to 62.11 and KK from 30.14 to 50.14. These consistent gains demonstrate that sample-level block-conditioned training effectively mitigates domain block size conflict and enables more suitable rollout structures for heterogeneous training samples. Overall, the results verify that Block-R1 provides a \textbf{stronger multi-domain RL framework for dLLMs by aligning each sample with its most informative decoding granularity during post-training.} More results are in Appendix~\ref{app:exp}.


\begin{figure*}[t]
\centering
\includegraphics[width=\linewidth]{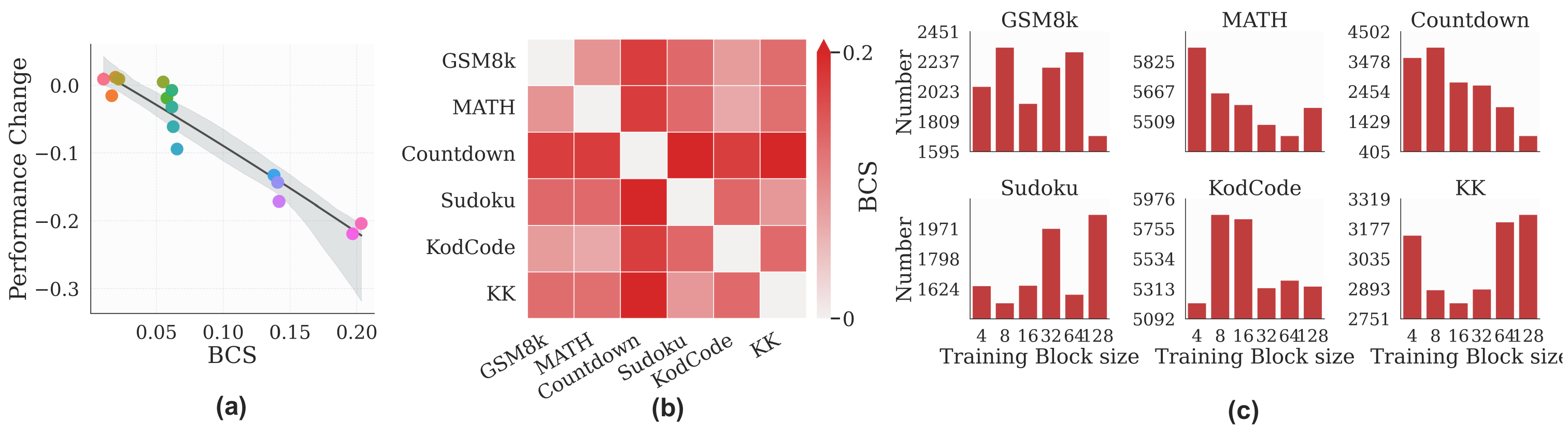}
\vspace{-0.7cm}
\caption{\textbf{Additional visualisations for domain block size conflict in multi-domain RL for dLLMs.}
\textbf{(a) Relationship between BCS and multi-domain RL performance}, where each point denotes domain pairs under vanilla fixed-block mix-domain RL and a larger BCS relates to stronger performance degradation.
\textbf{(b) Pairwise domain block size conflict visualisation}, where darker red cells indicate stronger block size conflict between two domains measured by BCS.
\textbf{(c) Sample-level best-improved training block size distribution}, where each bar counts the number of samples whose teacher-student improvement score is maximised by the corresponding block size group.}
\label{fig:block_conflict_all}
\vspace{-0.5cm}
\end{figure*}

\subsection{Insights into BCS and Multi-domain RL Post-training Performance}
\label{sec:bcs}
To examine whether the proposed BCS reflects the difficulty of multi-domain RL post-training, we visualise its relationship with fixed-block mix-domain RL performance in Figure~\ref{fig:block_conflict_all}(a). Each point denotes one domain pair jointly trained under the vanilla fixed-block setting, with colours detailed in Figure~\ref{fig:bcs_detail} in the Appendix. The x-axis denotes the BCS between two domains, while the y-axis denotes the mean performance change between single-domain RL to mix-domain RL over the two domains. As shown in Figure~\ref{fig:block_conflict_all}(a), larger BCS values generally lead to more severe performance degradation under vanilla multi-domain RL. This indicates that stronger disagreement in best-improved block size preferences makes it harder for a single fixed block size to provide suitable rollout structures for both domains. Therefore, \textbf{BCS serves as an important diagnostic indicator of multi-domain RL post-training effectiveness for dLLMs}.

\subsection{Visualisation of Domain Block Size Conflict}
\label{sec:heatmap}
We further visualise the pairwise BCS values across different reasoning domains in Figure~\ref{fig:block_conflict_all}(b). A larger BCS indicates that two domains prefer more divergent training block size distributions, suggesting that a unified fixed block size is less suitable for their joint RL post-training. Countdown exhibits clear block size conflicts with most other training domains, indicating that its preferred decoding granularity differs substantially from other reasoning tasks. This observation further supports the motivation of Block-R1. Instead of forcing Countdown and other domains to share the same fixed block size during multi-domain RL, Block-R1 assigns \textbf{sample-level best-improved training block sizes to mitigate such conflict. As a result, Block-R1 achieves remarkable performance improvement, especially on Countdown, as shown in Table~\ref{tab:full_cross_domain_llada}.} More insights and visualisations are provided in Appendix~\ref{app:best_block}.

\subsection{Block Size Distribution across Domains}
\label{sec:block_count}
The domain-level block size preference is further visualised in Figure~\ref{fig:block_conflict_all}(c). The x-axis denotes candidate block sizes, and the y-axis denotes sample counts. Tied block sizes are all counted. Following Equation~\ref{eq:improvement_and_best-improved_c}, we select the block size that maximises the teacher-student improvement score for each training sample, and then count the selected block sizes within each domain. The results show that different domains exhibit clearly different sample-level block size preferences. For example, Countdown favours smaller block sizes, while Sudoku and KK assign more samples to larger block sizes. This indicates that \textbf{the preferred block-based decoding granularity varies substantially across domains during dLLM post-training, further confirming the existence of domain block size conflict in multi-domain RL post-training for dLLMs.}

\begin{wraptable}{r}{0.5\textwidth}
\centering
\vspace{-1cm}
\caption{\textbf{Block-R1 with dynamic block size inference.}
\textit{b1} is applied at inference time after multi-domain RL post-training in Table~\ref{tab:full_cross_domain_llada}. Oracle denotes the performance by the best-performing block size for each test sample.}
\label{tab:block_r1_b1}
\resizebox{0.5\textwidth}{!}{
\begin{tabular}{l|cccc}
\toprule
\textbf{Method} & \textbf{GSM8K} & \textbf{MATH500} & \textbf{KodCode} & \textbf{KK} \\
\midrule
\textbf{Vanilla} & 57.24 & 28.20 & 22.60 & 30.14 \\
\textbf{Vanilla + \textit{b1}} & 59.82 & 30.20 & 23.20 & 38.86 \\
\textbf{Block-R1} & 80.74 & 35.80 & 28.60 & 50.14 \\
\textbf{Block-R1 + \textit{b1}} & 83.70 & 36.40 & 29.20 & 53.29 \\
\midrule
$\Delta_{\textit{b1}}$ over Block-R1 & +1.67 & +0.80 & +0.60 & +3.15 \\
\midrule
Oracle & 84.31 & 37.80 & 32.40 & 59.93 \\
\bottomrule
\end{tabular}
}
\vspace{-0.3cm}
\end{wraptable}
\subsection{Guidance for Test-time Development}
\label{sec:b1}
Block-R1 assigns each training sample its suitable decoding granularity during RL post-training, while remaining plug-and-play with dynamic block size methods at inference time. Therefore, we further combine the post-trained model with \textit{b1} to enable adaptive block size inference for test samples. As shown in Table~\ref{tab:block_r1_b1}, applying \textit{b1} consistently improves both Vanilla and Block-R1, and Block-R1 + \textit{b1} achieves much stronger performance than vanilla multi-domain RL. This suggests that Block-R1 can be freely plugged into existing dynamic block size methods that focus on the inference stage.

The Oracle results further highlight the value of Block-R1. Oracle selects the best-performing block size for each test sample generation, providing an empirical upper bound for sample-level test-time block size selection and showing that \textbf{the core idea of sample-level block allocation in Block-R1 is not only effective during RL post-training but also valuable at inference time.} The remaining gap between Block-R1 + \textit{b1} and Oracle indicates that stronger test-time block allocation methods can still bring further gains. Therefore, Block-R1 not only supports block-conditioned multi-domain RL training, \textbf{but also guides future dLLM test-time generation methods towards sample-level block size selection rather than globally fixed-size decoding across domains.}

\vspace{-0.3cm}
\section{Related Work}
\vspace{-0.3cm}
This work mainly relates to three research domains. \textbf{(1) Diffusion Large Language Models.} dLLMs provide an alternative generation paradigm to autoregressive language models by iteratively denoising masked or corrupted tokens~\cite{sohl2015deep, campbell2022continuous, meng2022concrete, austin2023structureddenoisingdiffusionmodels, lou2024discretediffusionmodelingestimating, shi2025simplifiedgeneralizedmaskeddiffusion, sahoo2024simpleeffectivemaskeddiffusion, nie2024scaling, zhu2025llada, dream2025, nie2025large, bie2025llada20, bie2026llada21}. Most dLLMs adopt block-based semi-autoregressive generation~\cite{2025blockdiffusion}. \textbf{(2) Reinforcement Learning for dLLM Reasoning.} Recent RL methods for dLLMs include token-level methods such as \textit{d1}~\cite{zhao2025d1}, \textit{d2}~\cite{wang2025d2}, and \textit{wd1}~\cite{tang2025wd1}, as well as sequence-level methods such as GDPO~\cite{rojas2025gdpo}, MDPO~\cite{he2025mdpo}, ESPO~\cite{ou2025espo}, SPG~\cite{wang2026spg}, StableDRL~\cite{zhong2026stabilizing}, DiRL~\cite{zhu2026DiRL}, and DARE~\cite{DARE}. Despite their progress, these methods generally use a fixed block size for post-training in a single domain scenario. \textbf{(3) Block-based Dynamic Inference} Recent inference-time methods improve dLLM decoding by adapting or accelerating the inference process~\cite{lu2025adablock, shu2026deferredcommitment, ma2025dinfer, huang2025ctrldiff, lee2025testtimescaling, b1, wang2026commit}. However, these methods mainly focus on inference-time decoding for dLLMs in single domain settings. In contrast, Block-R1 focuses on multi-domain RL post-training by assigning sample-level best-improved training block sizes during the post-training process. More detailed related work is provided in Appendix~\ref{app:related} due to page limit.

\vspace{-0.3cm}
\section{Conclusion}
\vspace{-0.3cm}
In this paper, we propose \textbf{Block-R1}, a novel benchmark for multi-domain RL post-training for dLLMs. Block-R1 formulates domain block size conflict and shows that a fixed one-for-all block size limits multi-domain RL. To address this, Block-R1 constructs a 41K block-based dataset where each sample is assigned a best-improved training block size. It further enables sample-level block-conditioned training without changing existing RL objectives. Experiments across 13 datasets, 7 RL algorithms, and diverse dLLMs demonstrate that Block-R1 improves dLLM cross-domain generalisation and provides a principled benchmark for guiding future test-time methods towards sample-level block size selection rather than globally fixed-size decoding for all tasks.
\bibliographystyle{abbrv}
\bibliography{neurips_2026}

\newpage
\appendix
\section{Appendix Overview}
This appendix provides supplementary materials, theoretical proofs, and comprehensive experimental details as follows:

\begin{itemize}[leftmargin=*, itemsep=2pt, topsep=2pt]
    \item \textbf{Appendix~\ref{app:related}} reviews detailed related literature, including:
    \begin{itemize}[leftmargin=*, label=$\circ$, itemsep=0pt]
        \item Diffusion Large Language Models in Appendix~\ref{app:dllm};
        \item Reinforcement Learning for LLM Reasoning in Appendix~\ref{app:reason};
        \item Reinforcement Learning for Block-based dLLM Reasoning in Appendix~\ref{app:rl};
        \item dLLMs' Block-based Generation Methods in Appendix~\ref{app:block_method}.
    \end{itemize}

    \item \textbf{Appendix~\ref{app:exp}} provides additional experimental analysis, including:
    \begin{itemize}[leftmargin=*, label=$\circ$, itemsep=0pt]
        \item Benchmarking multi-domain RL for different dLLMs in Appendix~\ref{app:multi_backbone};
        \item Benchmarking single-domain RL methods for dLLMs in Appendix~\ref{app:single_benchmark};
        \item Effectiveness of Block-R1 in single-domain RL in Appendix~\ref{app:single};
        \item Relationship between BCS and multi-domain RL performance in Appendix~\ref{app:bcs};
        \item Fixed-block conflict under Diffu-GRPO in Appendix~\ref{app:block_d1};
        \item Reward improvement under different training block sizes in Appendix~\ref{app:improve};
        \item Probability distribution of best-improved training block sizes in Appendix~\ref{app:best_block}.
    \end{itemize}

    \item \textbf{Appendix~\ref{app:reproduce}} provides detailed implementation and reproducibility settings, including:
    \begin{itemize}[leftmargin=*, label=$\circ$, itemsep=0pt]
        \item Detailed Dataset Settings and Statistic in Appendix~\ref{app:data};
        \item Supported dLLM Backbones in Appendix~\ref{app:backbone};
        \item Baseline Details in Appendix~\ref{app:baseline};
        \item Training Protocol in Appendix~\ref{app:training_protocol};
        \item Evaluation Protocol in Appendix~\ref{app:eval_protocol};
        \item Detailed Reward Functions in Appendix~\ref{app:reward_func};
        \item The computational complexity analysis in Appendix~\ref{app:complexity};
        \item Detailed Hyperparameter Settings in Table~\ref{tab:shared_defaults}.
    \end{itemize}

    \item \textbf{Appendix~\ref{sec:theoretical_proof}} presents theoretical and algorithmic details, including:
    \begin{itemize}[leftmargin=*, label=$\circ$, itemsep=0pt]
        \item The proof of Theorem~\ref{thm:conflict}, which establishes the sub-optimality of fixed-size block generation under domain block size conflict;
        \item The full Block-R1 dataset construction pipeline in Algorithm~\ref{app:algorithm};
    \end{itemize}

    \item \textbf{Appendix~\ref{app:limitation}} discusses the limitations of Block-R1.
\end{itemize}

\newpage
\section{Detailed Related Work}
\label{app:related}
\subsection{Diffusion Large Language Models}
\label{app:dllm}
Diffusion Large Language Models (dLLMs) provide an alternative generation paradigm to conventional autoregressive (AR) large language models. AR models generate text strictly from left to right by factorising the sequence probability into token-by-token conditional distributions. In contrast, dLLMs generate text through an iterative denoising process, where masked or corrupted tokens are progressively recovered conditioned on the surrounding context~\cite{sohl2015deep, campbell2022continuous, meng2022concrete, austin2023structureddenoisingdiffusionmodels, lou2024discretediffusionmodelingestimating, shi2025simplifiedgeneralizedmaskeddiffusion, sahoo2024simpleeffectivemaskeddiffusion, nie2024scaling, zhu2025llada, dream2025, nie2025large}. Recent advancements in dLLMs, such as LLaDA2.0~\cite{bie2025llada20} and LLaDA2.1~\cite{bie2026llada21}, further demonstrate the scalability and efficiency of dLLMs. Specifically, LLaDA2.0 scales dLLMs to 100B parameters through systematic AR-to-diffusion conversion and a three-phase block-level training scheme, while LLaDA2.1 accelerates diffusion decoding by integrating Token-to-Token editing into Mask-to-Token generation with a configurable threshold-decoding scheme. Other recent systems such as Dream~\cite{dream2025}, Mercury~\cite{labs2025mercuryultrafastlanguagemodels}, and LLaDA-style models~\cite{nie2025large} further show the potential of diffusion-based language modelling for efficient text generation.

Most existing dLLMs adhere to a semi-autoregressive generation paradigm with block-based generation strategies, where the output sequence is partitioned into blocks and tokens within each block are denoised in parallel~\cite{2025blockdiffusion}. This design improves generation parallelism while preserving sequential dependency across blocks in a global view. However, a fixed block size also imposes a rigid decoding granularity. In reasoning tasks, different samples and even different reasoning steps may require different amounts of context and different levels of intermediate verification. For example, arithmetic reasoning may benefit from smaller blocks that allow more fine-grained step-by-step control, while broader semantic or compositional reasoning may favour larger blocks that preserve long-range coherence~\cite{xiong2025minimalistapproachllmreasoning, ye2024beyond, tang2026multiplexthinking}. Therefore, treating block size as a static decoding hyperparameter can limit the ability of dLLMs to adapt their generation structure to the underlying reasoning process.

\subsection{Reinforcement Learning for LLM Reasoning}
\label{app:reason}
Reinforcement Learning (RL) has become an important post-training technique for improving the reasoning capability of large language models. Instead of only imitating reference answers, RL optimises model outputs according to task-level reward signals, such as mathematical correctness, code execution success, or format validity. Representative policy optimisation methods include TRPO~\cite{schulman2015trust}, PPO~\cite{schulman2017proximal}, and Group Relative Policy Optimisation (GRPO)~\cite{shao2024deepseekmath}. Compared with the standard PPO-style framework, GRPO simplifies optimisation by avoiding a separately trained value or critic model. Rather than estimating a value function, GRPO samples multiple outputs for the same prompt and computes relative advantages within the group, using the group reward statistics as a baseline.

Recent analyses suggest that many reasoning-oriented RL methods can be interpreted through weighted regression or selective fine-tuning perspectives~\cite{mroueh2025reinforcement, peng2019advantage, zhu2023fine, du2025simplifyrlhfrewardweightedsft, baheti2024leftoverlunchadvantagebasedoffline}. In particular, methods related to Rejection Sampling Fine-Tuning (RAFT)~\cite{xiong2025minimalistapproachllmreasoning} increase the likelihood of high-reward generations while reducing or ignoring low-reward generations. This connects RL-style reasoning optimisation to contrastive or noise-contrastive learning principles, where correct and incorrect generations provide different learning signals~\cite{zhu2025surprising, chen2025bridging, JMLR:v13:gutmann12a, oord2019representationlearningcontrastivepredictive}. Related preference optimisation methods, such as DPO and its variants~\cite{rafailov2023direct, rafailov2024directpreferenceoptimizationlanguage, azar2023generaltheoreticalparadigmunderstand, ethayarajh2024ktomodelalignmentprospect, wang2023reverseklgeneralizingdirect, hong2024orpomonolithicpreferenceoptimization}, also provide alternative perspectives on learning from reward or preference signals without explicit value-function estimation. These properties make group-based policy optimisation particularly suitable for reasoning tasks, where several valid solution paths may lead to the same correct answer.

\newpage
\begin{wrapfigure}{r}{0.6\textwidth}
\centering
\includegraphics[width=0.6\textwidth]{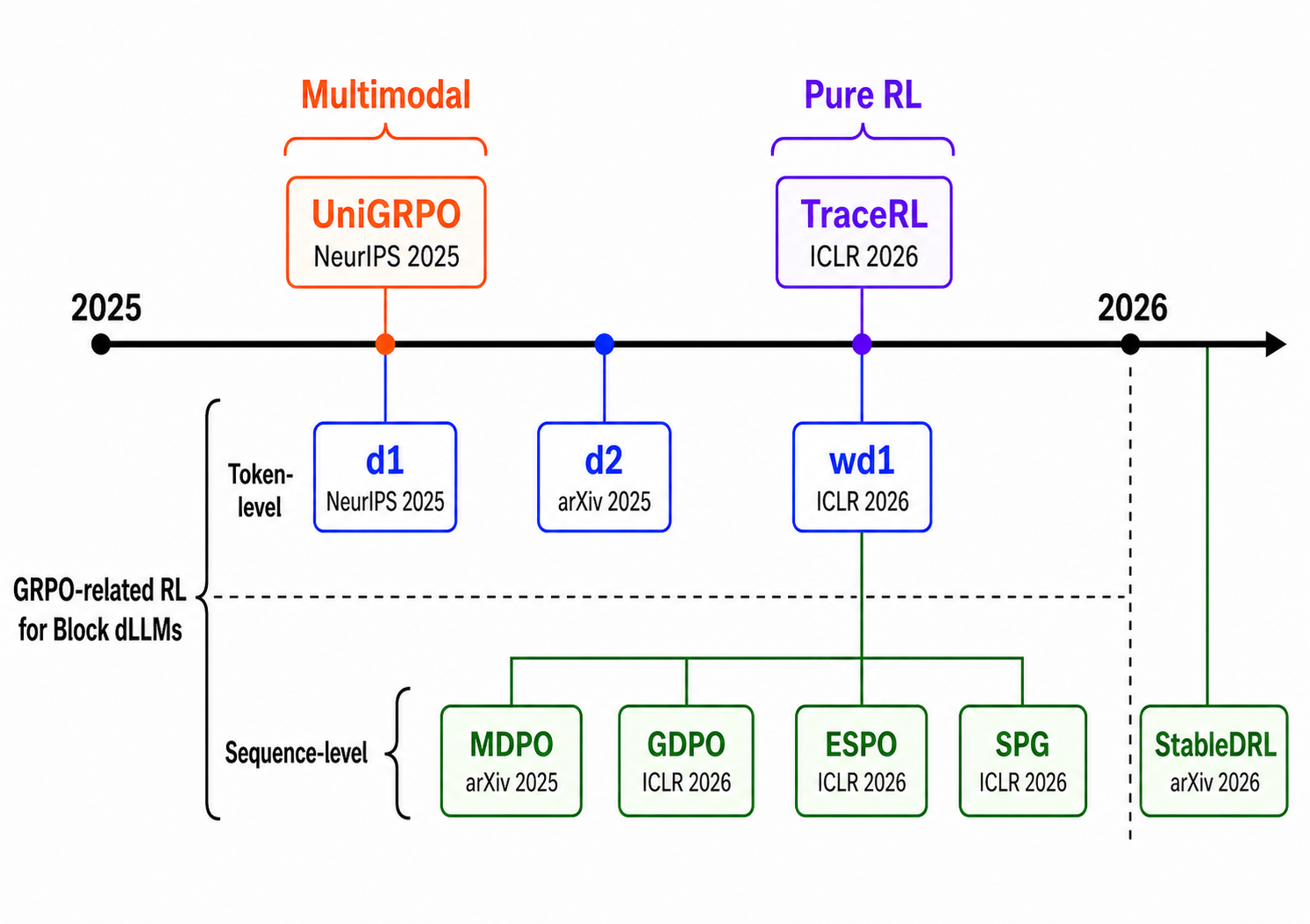}
\vspace{-0.7cm}
\caption{Development of RL methods for dLLMs. Existing methods can be organised into multimodal RL, pure RL, and GRPO-related RL. GRPO-related methods can be further grouped into token-level and sequence-level objectives.}
\vspace{-0.3cm}
\label{fig:RL}
\end{wrapfigure}
\subsection{Reinforcement Learning for Block-based dLLM Reasoning}
\label{app:rl}

Recent research on reinforcement learning (RL) for dLLMs can be broadly organised into three categories as illustrated in Figure~\ref{fig:RL}. The first category studies multimodal or diffusion-style RL, such as MMaDA~\cite{yang2025mmada}, diffusion policy optimisation~\cite{ding2024diffusion, wang2022diffusion, wang2024fine, ma2025soft, zhang2025energy, lu2023contrastive}, and related reward-guided diffusion optimisation~\cite{zekri2025fine, huang2025reinforcing}, where RL is applied to improve generation or reasoning under diffusion-style objectives. The second category explores pure RL frameworks for dLLMs, such as TraceRL~\cite{wang2025traceRL}, which attempts to optimise diffusion reasoning behaviour with conventional PPO-style policy optimisation. The third and most widely adopted category is GRPO-related RL for dLLMs, including \textit{d1}~\cite{zhao2025d1}, \textit{d2}~\cite{wang2025d2}, \textit{wd1}~\cite{tang2025wd1}, GDPO~\cite{rojas2025gdpo}, MDPO~\cite{he2025mdpo}, ESPO~\cite{ou2025espo}, SPG~\cite{wang2026spg}, StableDRL~\cite{zhong2026stabilizing}, DiRL~\cite{zhu2026DiRL}, and DARE~\cite{DARE}. These methods optimise dLLMs through group-relative or trajectory-level reward signals and can be further divided into token-level and sequence-level formulations depending on whether the policy surrogate is constructed over token-wise likelihood terms or over the full generated trajectory.

Token-level GRPO-related dLLM methods, such as \textit{d1}~\cite{zhao2025d1}, \textit{d2}~\cite{wang2025d2}, and \textit{wd1}~\cite{tang2025wd1}, optimise policies through token-level likelihood estimates or token-level likelihood reweighting. In \textit{d1} and related GRPO-style objectives, completion-level advantages are broadcast to token-wise likelihood ratios. In \textit{wd1}, token likelihoods are instead aggregated and reweighted through a ratio-free weighted objective. In both cases, rewards are computed at the completion level, but the optimisation signal is applied through token-level likelihood terms rather than a single sequence-level action. In contrast, sequence-level methods, such as MDPO~\cite{he2025mdpo}, GDPO~\cite{rojas2025gdpo}, ESPO~\cite{ou2025espo}, SPG~\cite{wang2026spg}, StableDRL~\cite{zhong2026stabilizing}, and DiRL~\cite{zhu2026DiRL}, treat the complete denoising trajectory or generated completion as the optimisation unit. These methods estimate a sequence-level likelihood or sequence-level policy objective, and then optimise the policy using trajectory-level reward signals.

Despite the progress of existing RL for dLLM methods, both token-level and sequence-level GRPO-related methods generally adopt a fixed block size during rollout generation and policy optimisation. This creates a critical gap between the block-based parallel generation structure of dLLMs and the RL policy optimisation objective, as the optimisation process does not explicitly model how block size reshapes the trajectory distribution used for rollout sampling and reward estimation across different tasks. As a result, block size is treated as a static decoding hyperparameter rather than a structural factor that can influence trajectory sampling, reward estimation, and policy updates. Overall, existing RL for dLLM methods generally neglect the importance of block-size adaptation during RL post-training for dLLMs, especially in cross-domain reasoning settings where different tasks may require different parallel decoding granularities.

\subsection{dLLMs' Block-based Generation Methods}
\label{app:block_method}

Recent studies have advanced block-based generation methods for dLLMs mainly from the inference-time perspective. AdaBlock-dLLM~\cite{lu2025adablock} is a tuning-free method that searches for the next newline character and truncates the current block at that position. Although it provides variable block sizes during inference, it relies on pre-defined confidence thresholds and hard-coded newline-based splitting, where newline characters may not always correspond to true reasoning-step boundaries. Deferred Commitment Decoding (DCD)~\cite{shu2026deferredcommitment} introduces a confidence-aware sliding window to defer high-uncertainty tokens until sufficient context is available. Other inference-oriented methods such as dInfer~\cite{ma2025dinfer}, CtrlDiff~\cite{huang2025ctrldiff}, and test-time scaling with hidden semi-autoregressive experts~\cite{lee2025testtimescaling} further improve dLLM decoding efficiency or controllability. Moreover, VSB~\cite{wang2026commit} enables more reliable blockwise decoding for dLLMs by selecting variable-size block boundaries based on whether current token predictions remain consistent with or without access to future context. These methods improve the inference process of a fixed pre-trained dLLM, but they do not modify the post-training stage. Therefore, a gap remains between the fixed-block rollout structure used during dLLM post-training and the adaptive decoding strategy used during inference. In addition, \textit{b1}~\cite{b1} enables dLLMs to generate dynamic-size reasoning blocks by dynamically aligning reasoning-step boundaries with block boundaries to improve dLLM reasoning quality.

However, these methods mainly focus on improving the inference behaviour of a fixed or post-trained dLLM. In contrast, Block-R1 focuses on improving the cross-domain generalisability of dLLMs at post-training time via adaptive block size. Specifically, Block-R1 assigns sample-level training block sizes during multi-domain RL rollout generation, thereby addressing domain block size conflict before inference. Moreover, Block-R1 remains compatible with inference-time dynamic block size methods, allowing it to be further combined with methods such as \textit{b1} for additional improvements.

\newpage
\section{Additional Experiments}
\label{app:exp}
This section provides additional empirical analysis for Block-R1. It first evaluates the generality of Block-R1 across different dLLM backbones in Section~\ref{app:multi_backbone}. Then, it benchmarks existing single-domain RL methods for dLLMs in Section~\ref{app:single_benchmark} and evaluates the effectiveness of Block-R1 in single-domain RL settings in Section~\ref{app:single}. We further show the relationship between Block Size Conflict Score (BCS) and multi-domain RL performance in Section~\ref{app:bcs}, support the fixed-block conflict claim under another representative dLLM RL algorithm in Section~\ref{app:block_d1}, and visualise how different domains exhibit different reward improvements and best-improved training block size preferences in Sections~\ref{app:improve} and~\ref{app:best_block}.

\begin{table*}[t]
\centering
\caption{\textbf{Benchmarking multi-domain RL post-training of different dLLMs.}
Each backbone is post-trained on \textbf{all training domains} in Table~\ref{tab:dataset_stats} to ensure the same amount of training data. The fixed-block baseline uses a block size of 32, and the default generation length is 256. \textbf{+ Block-R1} denotes sample-level block-conditioned multi-domain RL by our dataset. The best checkpoint accuracies are reported under a consistent zero-shot setting and are highlighted by {\color{purple}\textbf{best}} and {\color{teal}\underline{runner-up}}.}
\label{tab:multi_backbone_benchmark}
\vspace{-0.2cm}
\resizebox{\linewidth}{!}{
\begin{tabular}{l|ccccccccccc}
\toprule
\textbf{Model} 
& \textbf{Countdown} & \textbf{GSM8K} & \textbf{MATH500} & \textbf{KodCode} & \textbf{Sudoku} & \textbf{KK} & \textbf{HellaSwag} & \textbf{MMLU} & \textbf{ARC-E} & \textbf{MMLU-Pro} & \textbf{ARC-C} \\
\midrule
GSAI-ML/LLaDA-8B-Instruct & 30.08 & 57.24 & 28.20 & 22.60 & 9.77 & 30.14 & 52.10 & 52.06 & 73.95 & 29.79 & 65.87 \\
\quad + Block-R1 & 62.11 & 80.74 & 35.80 & {\color{teal}\underline{28.60}} & {\color{teal}\underline{26.95}} & {\color{teal}\underline{50.14}} & 64.07 & 62.22 & {\color{purple}\textbf{90.53}} & {\color{teal}\underline{37.96}} & {\color{teal}\underline{82.51}} \\
\quad $\Delta$ & +38.28 & +24.79 & +7.40 & +6.00 & +17.18 & +20.00 & +11.97 & +10.16 & +16.58 & +8.17 & +16.64 \\

\midrule
GSAI-ML/LLaDA-1.5 & 25.39 & 66.41 & 24.40 & 15.80 & 12.27 & 35.57 & 50.56 & 55.42 & 75.59 & 30.49 & 63.31 \\
\quad + Block-R1 & {\color{teal}\underline{58.98}} & 81.88 & 36.00 & 28.40 & 24.48 & 48.57 & 58.89 & {\color{purple}\textbf{64.36}} & {\color{teal}\underline{88.93}} & 35.44 & 77.99 \\
\quad $\Delta$ & +33.59 & +15.47 & +11.60 & +12.60 & +12.21 & +13.00 & +8.33 & +8.94 & +13.34 & +4.95 & +14.68 \\

\midrule
Dream-org/Dream-Instruct-7B & 19.92 & 62.09 & 30.20 & 15.20 & 10.04 & 28.86 & 54.37 & 53.72 & 77.65 & 27.84 & 60.41 \\
\quad + Block-R1 & 53.52 & 82.26 & 35.20 & 27.60 & 22.98 & 49.14 & 62.25 & 58.79 & 85.52 & 33.99 & 75.85 \\
\quad $\Delta$ & +33.60 & +20.17 & +5.00 & +12.40 & +12.94 & +20.28 & +7.88 & +5.07 & +7.87 & +6.15 & +15.44 \\

\midrule
JetLM/SDAR-8B-Chat-b32 & 24.22 & 70.36 & 25.40 & 19.80 & 10.61 & 32.29 & 49.96 & 52.25 & 79.34 & 32.28 & 68.94 \\
\quad + Block-R1 & 55.47 & {\color{teal}\underline{84.53}} & {\color{teal}\underline{36.60}} & 25.80 & 24.19 & 45.57 & {\color{teal}\underline{69.98}} & 59.42 & 88.51 & 35.37 & 79.78 \\
\quad $\Delta$ & +31.25 & +14.17 & +11.20 & +6.00 & +13.58 & +13.28 & +20.02 & +7.17 & +9.17 & +3.09 & +10.84 \\

\midrule
Gen-Verse/TraDo-8B-Instruct & 26.95 & 58.83 & 23.60 & 21.20 & 14.33 & 30.14 & 46.40 & 49.58 & 80.72 & 28.69 & 65.70 \\
\quad + Block-R1 & 50.00 & 81.27 & 34.80 & 24.40 & 20.68 & 43.43 & 66.32 & 60.67 & 79.97 & 32.25 & 77.65 \\
\quad $\Delta$ & +23.05 & +22.44 & +11.20 & +3.20 & +6.35 & +13.29 & +19.92 & +11.09 & -0.75 & +3.56 & +11.95 \\

\midrule
inclusionAI/LLaDA2.0-16B & 39.06 & 70.81 & 31.40 & 24.20 & 19.60 & 42.43 & 55.10 & 58.85 & 82.28 & 32.42 & 70.65 \\
\quad + Block-R1 & {\color{purple}\textbf{64.45}} & {\color{purple}\textbf{85.52}} & {\color{purple}\textbf{42.80}} & {\color{purple}\textbf{31.00}} & {\color{purple}\textbf{28.66}} & {\color{purple}\textbf{54.57}} & {\color{purple}\textbf{74.62}} & {\color{teal}\underline{63.64}} & 85.69 & {\color{purple}\textbf{38.58}} & {\color{purple}\textbf{84.73}} \\
\quad $\Delta$ & +25.39 & +14.71 & +11.40 & +6.80 & +9.06 & +12.14 & +19.52 & +4.79 & +3.41 & +6.16 & +14.08 \\

\bottomrule
\end{tabular}
}
\vspace{-0.3cm}
\end{table*}
\subsection{Multi-domain Benchmark across dLLM Backbones}
\label{app:multi_backbone}

To further evaluate whether Block-R1 can generalise beyond a single backbone, we construct a multi-domain benchmark across representative dLLM backbones. StableDRL is adopted as the default RL algorithm due to its strong empirical performance. For the fixed-block multi-domain baseline, the block size is set to 32, and the default generation length is set to 256. The benchmark covers mathematical reasoning, code generation, puzzle solving, general capabilities, and advanced reasoning tasks.



\begin{table*}[ht]
\centering
\caption{\textbf{Benchmarking single-domain RL methods for dLLMs.} \colorbox{academicBlue!20}{Blue cells} denote base performance without RL. \colorbox{orange!20}{Orange cells} denote in-domain RL where training and testing domains are the same. Test accuracy (\%) is highlighted by {\color{purple}\textbf{best}} and {\color{teal}\underline{runner-up}}.}
\label{table:single_benchmark}
\resizebox{\linewidth}{!}{%
\begin{tabular}{l|cc|cc|cc|cc|cc|cc|cc|cc}
\toprule
\multirow{2}{*}{\textbf{Methods}} 
& \multicolumn{2}{c|}{\textbf{Countdown}} & \multicolumn{2}{c|}{\textbf{GSM8K}} & \multicolumn{2}{c|}{\textbf{MATH500}} & \multicolumn{2}{c|}{\textbf{HumanEval}} & \multicolumn{2}{c|}{\textbf{MBPP}} & \multicolumn{2}{c|}{\textbf{KodCode}} & \multicolumn{2}{c|}{\textbf{Sudoku}} & \multicolumn{2}{c}{\textbf{KK}} \\
& \textbf{256} & \textbf{512} & \textbf{256} & \textbf{512} & \textbf{256} & \textbf{512} & \textbf{256} & \textbf{512} & \textbf{256} & \textbf{512} & \textbf{256} & \textbf{512} & \textbf{256} & \textbf{512} & \textbf{256} & \textbf{512} \\

\midrule
\rowcolor{academicBlue!20}
LLaDA-8B-Instruct & 16.80 & 16.02 & 76.19 & 77.63 & 32.00 & 34.60 & 28.66 & 35.37 & 33.60 & 40.40 & 22.00 & 26.20 & 7.81 & 8.20 & 30.00 & 37.29 \\
\rowcolor{academicBlue!20}
+ AdaBlock-dLLM & 14.84 & 16.80 & 76.27 & 77.94 & 32.40 & 33.60 & 30.49 & 37.20 & 34.00 & 39.60 & 22.40 & 24.80 & 6.64 & 7.81 & 33.71 & 37.43 \\
\rowcolor{academicBlue!20}
+ SFT & 15.23 & 20.70 & 75.59 & 78.54 & 32.20 & 34.40 & 28.05 & 35.98 & 30.40 & 41.00 & 19.40 & 25.80 & 8.20 & 7.03 & 32.00 & 36.57 \\

\midrule
\rowcolor{orange!20}
+ \textit{Diffu}-GRPO & 19.92 & 24.22 & 76.35 & 79.98 & 33.60 & 36.80 & 29.27 & 36.59 & 34.00 & 40.80 & 24.60 & 26.40 & 13.67 & 12.11 & 36.86 & 37.43 \\
\rowcolor{orange!20}
+ \textit{d1} & 25.39 & 26.17 & 77.03 & 80.67 & 33.40 & {\color{purple}\textbf{37.80}} & 29.88 & 36.59 & 34.40 & 40.80 & 25.20 & 26.60 & 15.23 & 18.36 & 37.00 & 37.57 \\
\rowcolor{orange!20}
+ GDPO & 17.97 & 20.70 & 76.57 & 79.23 & 32.20 & 35.00 & 29.27 & 37.80 & {\color{purple}\textbf{35.00}} & 41.60 & {\color{teal}\underline{26.80}} & {\color{purple}\textbf{28.00}} & 9.77 & 10.55 & 36.86 & 38.86 \\
\rowcolor{orange!20}
+ ESPO & {\color{teal}\underline{56.64}} & {\color{teal}\underline{52.34}} & 80.21 & 79.76 & {\color{teal}\underline{35.00}} & {\color{teal}\underline{37.60}} & {\color{purple}\textbf{31.71}} & {\color{teal}\underline{40.85}} & 33.80 & 40.60 & 25.80 & 26.40 & 20.31 & 19.53 & 36.14 & 37.29 \\
\rowcolor{orange!20}
+ SPG & 47.27 & 44.53 & {\color{teal}\underline{81.05}} & 80.29 & 34.80 & 35.20 & 30.49 & 39.63 & 34.20 & 41.40 & 26.40 & 27.00 & {\color{teal}\underline{23.44}} & {\color{purple}\textbf{23.83}} & 36.57 & 38.29 \\
\rowcolor{orange!20}
+ \textit{wd1} & 39.45 & 38.67 & 78.85 & {\color{purple}\textbf{81.65}} & 34.20 & 37.40 & 30.49 & 37.80 & 34.00 & {\color{teal}\underline{41.80}} & 25.60 & 27.40 & 23.05 & {\color{teal}\underline{23.05}} & {\color{purple}\textbf{38.00}} & {\color{purple}\textbf{39.71}} \\
\rowcolor{orange!20}
+ StableDRL & {\color{purple}\textbf{58.98}} & {\color{purple}\textbf{62.11}} & {\color{purple}\textbf{81.35}} & {\color{teal}\underline{81.05}} & {\color{purple}\textbf{35.40}} & {\color{teal}\underline{37.60}} & {\color{teal}\underline{31.10}} & {\color{purple}\textbf{43.29}} & {\color{teal}\underline{34.80}} & {\color{purple}\textbf{42.40}} & {\color{purple}\textbf{28.20}} & {\color{teal}\underline{27.60}} & {\color{purple}\textbf{25.39}} & {\color{purple}\textbf{23.83}} & {\color{teal}\underline{37.57}} & {\color{teal}\underline{39.29}} \\
\bottomrule
\end{tabular}
}
\end{table*}
\subsection{Benchmarking Single-domain RL Methods for dLLMs}
\label{app:single_benchmark}

Table~\ref{table:single_benchmark} provides a comprehensive single-domain benchmark for various latest RL methods on dLLMs. The results show that RL post-training generally improves over the base model across different reasoning domains, yet the improvement varies substantially across algorithms and tasks. In particular, StableDRL achieves the strongest overall performance on most domains, including Countdown, GSM8K, MATH500, HumanEval, MBPP, KodCode, and Sudoku, while \textit{wd1} performs competitively on KK. This suggests that recent dLLM RL methods can effectively improve reasoning performance, but their gains remain domain dependent.

Also, different reasoning tasks benefit from different RL algorithms. For example, ESPO and StableDRL produce large improvements on Countdown, while \textit{wd1} and StableDRL are more competitive on KK. In contrast, some domains such as MATH500 and KodCode show relatively smaller gains, indicating that single-domain RL does not uniformly improve all reasoning tasks. These results highlight the need to improve the dLLM generalisability across all domains beyond the single domain scenarios.

\begin{table*}[ht]
\centering
\caption{\textbf{Evaluation of different RL methods for dLLMs in the single-domain setting.}
The Block-R1-41K dataset is restricted to single-domain RL to compare with different RL methods using the same amount of training data. The best checkpoint accuracies are reported under a consistent zero-shot setting and highlighted by {\color{purple}\textbf{best}} and {\color{teal}\underline{runner-up}}.}
\label{table:block_r1_single_domain}
\vspace{-0.2cm}
\resizebox{0.9\linewidth}{!}{%
\begin{tabular}{l|cc|cc|cc|cc|cc|cc}
\toprule
\multirow{2}{*}{\textbf{Methods}} & \multicolumn{2}{c|}{\textbf{Countdown}} & \multicolumn{2}{c|}{\textbf{GSM8K}} & \multicolumn{2}{c|}{\textbf{MATH500}} & \multicolumn{2}{c|}{\textbf{KodCode}} & \multicolumn{2}{c|}{\textbf{Sudoku}} & \multicolumn{2}{c}{\textbf{KK}} \\& \textbf{256} & \textbf{512} & \textbf{256} & \textbf{512} & \textbf{256} & \textbf{512} & \textbf{256} & \textbf{512} & \textbf{256} & \textbf{512} & \textbf{256} & \textbf{512} \\
\midrule
\rowcolor{academicBlue!20}
LLaDA-8B-Instruct & 16.80 & 16.02 & 76.19 & 77.63 & 32.00 & 34.60 & 22.00 & 26.20 & 7.81 & 8.20 & 30.00 & 37.29 \\
\rowcolor{academicBlue!20}
+ SFT & 15.23 & 20.70 & 75.59 & 78.54 & 32.20 & 34.40 & 19.40 & 25.80 & 8.20 & 7.03 & 32.00 & 36.57 \\
\midrule
\rowcolor{orange!20}
+ \textit{Diffu}-GRPO & 19.92 & 24.22 & 76.35 & 79.98 & 33.60 & 36.80 & 24.60 & 26.40 & 13.67 & 12.11 & 36.86 & 37.43 \\
\rowcolor{orange!20}
+ \textit{Diffu}-GRPO (Block-R1) & 26.17 & 31.25 & 79.68 & 80.36 & 34.00 & 37.20 & 26.20 & 27.00 & 15.85 & 16.40 & 37.00 & 37.71 \\
\rowcolor{orange!20}
\quad $\Delta$ & +6.25 & +7.03 & +3.33 & +0.38 & +0.40 & +0.40 & +1.60 & +0.60 & +2.18 & +4.29 & +0.14 & +0.28 \\
\midrule
\rowcolor{orange!20}
+ \textit{wd1} & 39.45 & 38.67 & 78.85 & 81.65 & 34.20 & 37.40 & 25.60 & 27.40 & 23.05 & 23.05 & 38.00 & 39.71 \\
\rowcolor{orange!20}
+ \textit{wd1} (Block-R1) & 42.58 & 43.36 & {\color{purple}\textbf{82.41}} & {\color{teal}\underline{82.71}} & 36.40 & {\color{purple}\textbf{37.80}} & 25.80 & {\color{teal}\underline{28.00}} & 23.78 & {\color{teal}\underline{24.49}} & {\color{teal}\underline{44.57}} & {\color{teal}\underline{45.86}} \\
\rowcolor{orange!20}
\quad $\Delta$ & +3.13 & +4.69 & +3.56 & +1.06 & +2.20 & +0.40 & +0.20 & +0.60 & +0.73 & +1.44 & +6.57 & +6.15 \\
\midrule
\rowcolor{orange!20}
+ StableDRL & {\color{teal}\underline{58.98}} & {\color{teal}\underline{62.11}} & 81.35 & 81.05 & 35.40 & {\color{teal}\underline{37.60}} & {\color{teal}\underline{28.20}} & 27.60 & {\color{teal}\underline{25.39}} & 23.83 & 37.57 & 39.29 \\
\rowcolor{orange!20}
+ \textit{StableDRL} (Block-R1) & {\color{purple}\textbf{60.16}} & {\color{purple}\textbf{64.06}} & {\color{teal}\underline{82.34}} & {\color{purple}\textbf{83.47}} & {\color{purple}\textbf{37.00}} & {\color{purple}\textbf{37.80}} & {\color{purple}\textbf{29.00}} & {\color{purple}\textbf{28.80}} & {\color{purple}\textbf{26.35}} & {\color{purple}\textbf{25.15}} & {\color{purple}\textbf{47.57}} & {\color{purple}\textbf{49.57}} \\
\rowcolor{orange!20}
\quad $\Delta$ & +1.18 & +1.95 & +0.99 & +2.42 & +1.60 & +0.20 & +0.80 & +1.20 & +0.96 & +1.32 & +10.00 & +10.28 \\
\bottomrule
\end{tabular}%
}
\end{table*}
\subsection{Effectiveness of Block-R1 in Single-Domain RL}
\label{app:single}
Block-R1 remains effective in single-domain RL scenarios. As shown in Table~\ref{table:block_r1_single_domain}, applying Block-R1 consistently improves performance over the corresponding single-domain RL baseline on most reasoning domains and generation lengths over various existing RL algorithms. The gains are especially clear on Countdown, GSM8K, and KK, indicating that block-size selection remains important even when training and testing are performed within the same domain. This can be attributed to the fine-grained sample-level best-improved training block size adaptation in Block-R1, which allows different training samples to use suitable decoding granularities rather than enforcing a fixed block size for the entire domain.

\begin{table*}[h]
\centering
\caption{\textbf{Relationship between Block Size Conflict Score (BCS) and dLLM Multi-domain RL performance.}
Domain pairs with larger BCS indicate stronger disagreement in best-improved training block sizes, while smaller BCS pairs indicate more compatible training block-size preferences. Mix denotes fixed-block RL training on both domains, while In-domain denotes post-training separately. Negative $\Delta$ indicates degradation from in-domain RL, while positive $\Delta$ indicates improvement.}
\label{tab:bcs}
\vspace{-0.2cm}
\resizebox{0.9\linewidth}{!}{
\begin{tabular}{lllccccc}
\toprule
\textbf{Group} & \textbf{Domain A} & \textbf{Domain B} & \textbf{BCS} & \textbf{A In-domain} & \textbf{A Mix ($\Delta$)} & \textbf{B In-domain} & \textbf{B Mix ($\Delta$)} \\
\midrule
\multirow{3}{*}{\textbf{Large BCS}} & Countdown & KK & 0.1969 & 58.98 & 23.44 \textcolor{red}{(-35.54)} & 37.57 & 29.29 \textcolor{red}{(-8.28)} \\
& Countdown & MATH500 & 0.1418 & 58.98 & 30.08 \textcolor{red}{(-28.90)} & 35.40 & 30.00 \textcolor{red}{(-5.40)} \\
& Countdown & KodCode & 0.1380 & 58.98 & 33.20 \textcolor{red}{(-25.78)} & 28.20 & 27.40 \textcolor{red}{(-0.80)} \\
\midrule
\multirow{3}{*}{\textbf{Small BCS}} & MATH500 & KK & 0.0551 & 35.40 & 35.80 \textcolor{teal}{(+0.40)} & 37.57 & 38.14 \textcolor{teal}{(+0.57)} \\
& MATH500 & GSM8K & 0.0219 & 35.40 & 36.00 \textcolor{teal}{(+0.60)} & 81.35 & 81.88 \textcolor{teal}{(+0.53)} \\
& MATH500 & KodCode & 0.0103 & 35.40 & 36.00 \textcolor{teal}{(+0.60)} & 28.20 & 30.40 \textcolor{teal}{(+2.20)} \\
\bottomrule
\end{tabular}
}
\vspace{-0.2cm}
\end{table*}
\subsection{How BCS Affects dLLM Multi-domain Post-training}
\label{app:bcs}
To evaluate whether block size conflict affects multi-domain RL process for dLLMs, we further compare domain pairs with large and small BCS values. As shown in Table~\ref{tab:bcs}, large-BCS pairs consistently suffer performance degradation under fixed-block mix-domain RL. For example, mixing Countdown with KK, MATH500, or KodCode substantially reduces Countdown performance compared with its in-domain RL result. In contrast, small-BCS pairs lead to stable or improved performance on both domains, suggesting that compatible block-size preferences make fixed-block mix-domain RL less harmful. These results show that BCS is closely associated with dLLM multi-domain RL post-training performance as larger BCS tends to cause stronger performance degradation under fixed-block mix-domain RL, while smaller BCS corresponds to more compatible performance.

\begin{figure*}[h]
    \vspace{-0.4cm}
    \centering
    \includegraphics[width=0.7\linewidth]{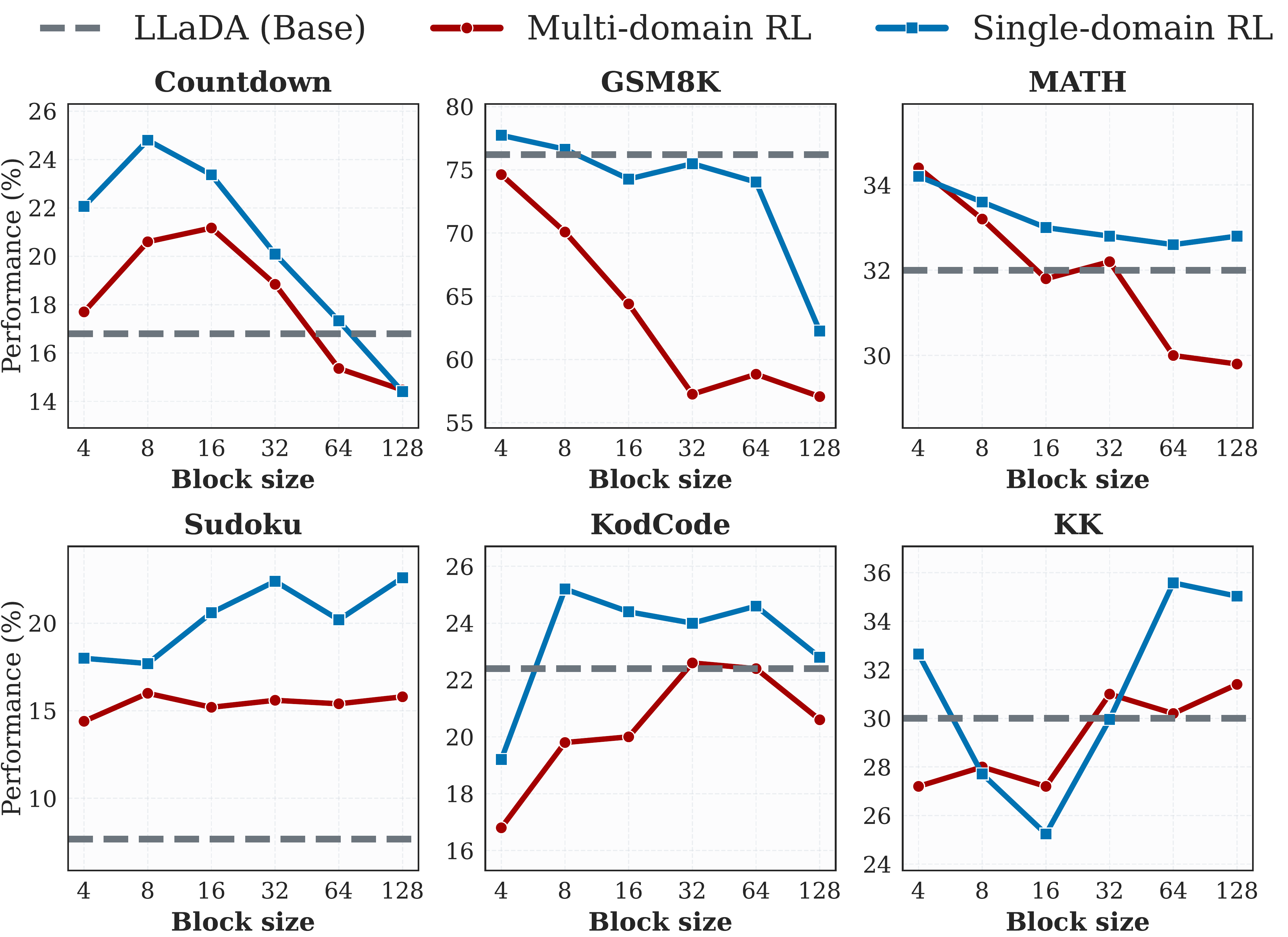}
    \vspace{-0.2cm}
    \caption{Motivation for Block-R1. Multi-domain RL refers to using all six domains as training domains during post-training. Diffu-GRPO~\cite{zhao2025d1}, a standard GRPO-based RL method for dLLMs, is used as the default base RL algorithm for comparison, while LLaDA (8B)~\cite{nie2025large} is adopted as the dLLM backbone following most existing dLLM post-training methods~\cite{zhao2025d1, tang2025wd1, rojas2025gdpo}. Two observations are highlighted: \textbf{(1) RL post-training on different domains favours different block sizes. (2) Multi-domain post-training on all domains with the same block size (red line) leads to degraded dLLM test performance compared with single-domain RL (blue line)}, and can even perform worse than the base model without RL (grey dashed line).}
    \vspace{-0.4cm}
    \label{fig:block_d1}
\end{figure*}
\subsection{Fixed-block Multi-domain Conflict for dLLMs RL Post-training}
\label{app:block_d1}
Figure~\ref{fig:block_d1} revisits the motivation of Block-R1 using Diffu-GRPO as the underlying RL algorithm. Similar to the observations of StableDRL in Figure~\ref{fig:block_stable}, different domains achieve their best single-domain RL performance under different block sizes, which indicates that the preferred parallel decoding granularity is inherently domain dependent. Therefore, a one-for-all block size is unlikely to provide a suitable rollout structure for all domains simultaneously.

More importantly, when all six domains are jointly used for RL post-training with one fixed block size, the resulting multi-domain performance consistently underperforms the corresponding single-domain RL performance across domains. In several cases, the degradation is so severe that the multi-domain RL model even performs worse than the base model without RL. \textbf{This result further confirms that the domain conflict in dLLM RL post-training by fixed-size block generation is severe and not specific to one particular RL optimisation algorithm.}

\begin{figure*}[t]
    \centering
    \includegraphics[width=0.7\linewidth]{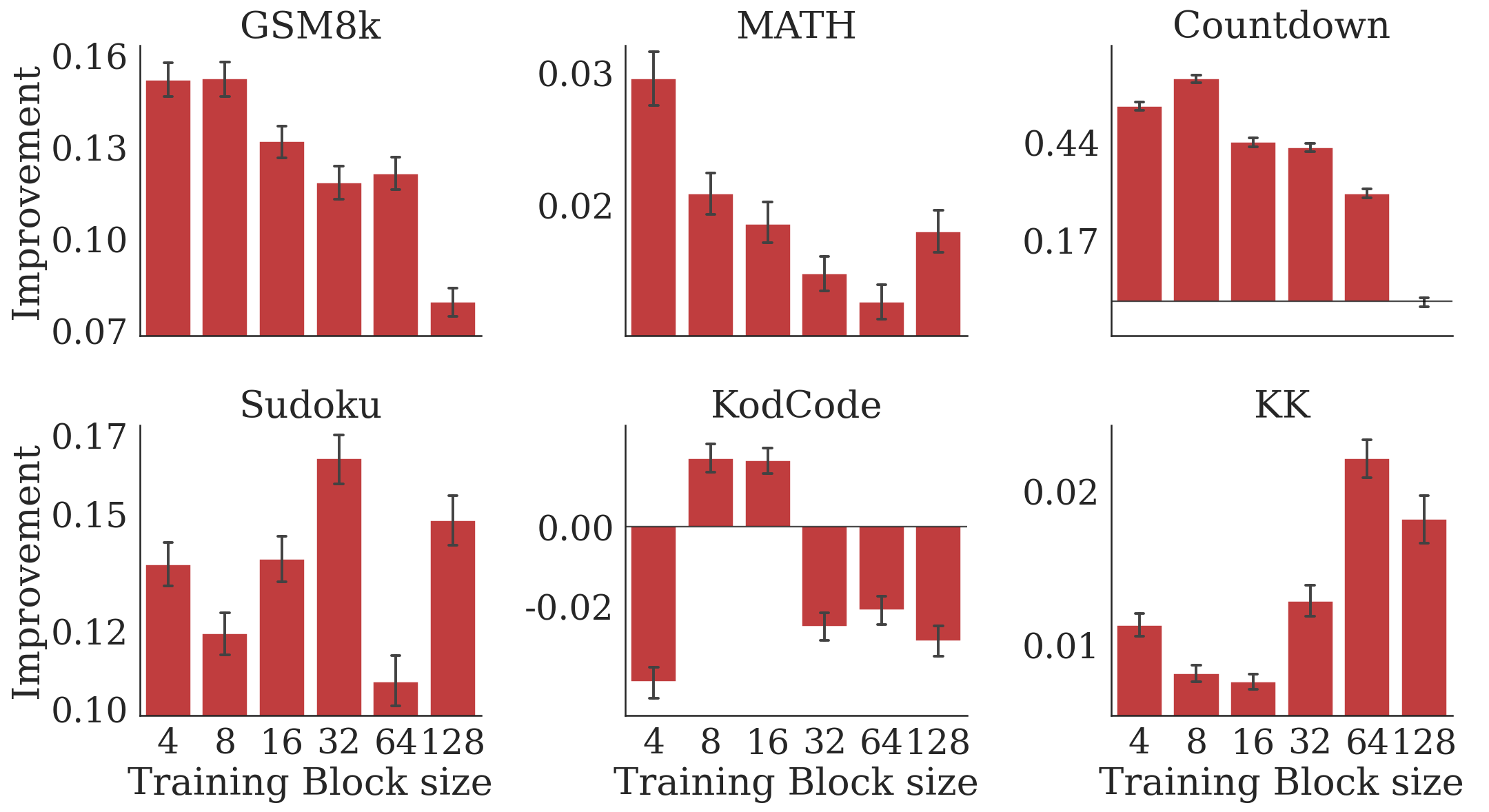}
    \caption{\textbf{Average reward improvement under different training block sizes.}
    For each domain and each block size $c$, the bar shows the mean teacher-student improvement $\mathbb{E}[\Delta(x,c)\mid x\sim\mathcal{D}_k]$, where $\Delta(x,c)=A_{\theta_T}(x,c)-A_{\theta_S}(x,c)$. Error bars denote 95\% confidence intervals. \textbf{The results show that block size significantly affects the reward improvement obtained during dLLM RL post-training across different domains.}}
    \label{fig:block_improve}
\end{figure*}
\subsection{Reward Improvement under Different Training Block Sizes}
\label{app:improve}
To further analyse the effect of block size on reward improvement, we visualise the average teacher-student improvement under different training block sizes for each domain. For each sample $x$ and block size $c$, the improvement score is computed as $\Delta(x,c)=A_{\theta_T}(x,c)-A_{\theta_S}(x,c)$, following Equation~\ref{eq:improvement_and_best-improved_c}. For each domain and each candidate block size, we aggregate the improvement scores over all samples from that domain and report the mean value $\mathbb{E}[\Delta(x,c)\mid x\sim\mathcal{D}_k]$. The error bars denote the 95\% confidence interval, reflecting the sample-level variation within each domain. Specifically, the x-axis represents the training block size, while the y-axis represents the average reward improvement. The horizontal zero line indicates the baseline where the teacher and student achieve no reward difference.

Intuitively, a positive value means that the teacher obtains a higher reward than the student under the corresponding block size, indicating reachable improvement space for RL post-training. As shown in Figure~\ref{fig:block_improve}, the reward improvement varies substantially across block sizes and domains. For example, mathematical reasoning and logical puzzle domains exhibit distinct preferred block-size regions. Some domains, such as KodCode, even exhibit negative reward improvement under certain block sizes, such as $64$ and $128$, indicating that generation under these block sizes provides limited or even harmful learning signals for dLLM RL post-training. Therefore, block size serves as an important factor for identifying whether a sample or domain can provide effective reward improvement under a specific dLLM block-based decoding structure. \textbf{This further suggests that block size is not a neutral decoding hyperparameter, but a key structural factor influencing reward estimation and RL post-training effectiveness for dLLMs.}


\begin{wrapfigure}{r}{0.4\textwidth}
\centering
\includegraphics[width=0.4\textwidth]{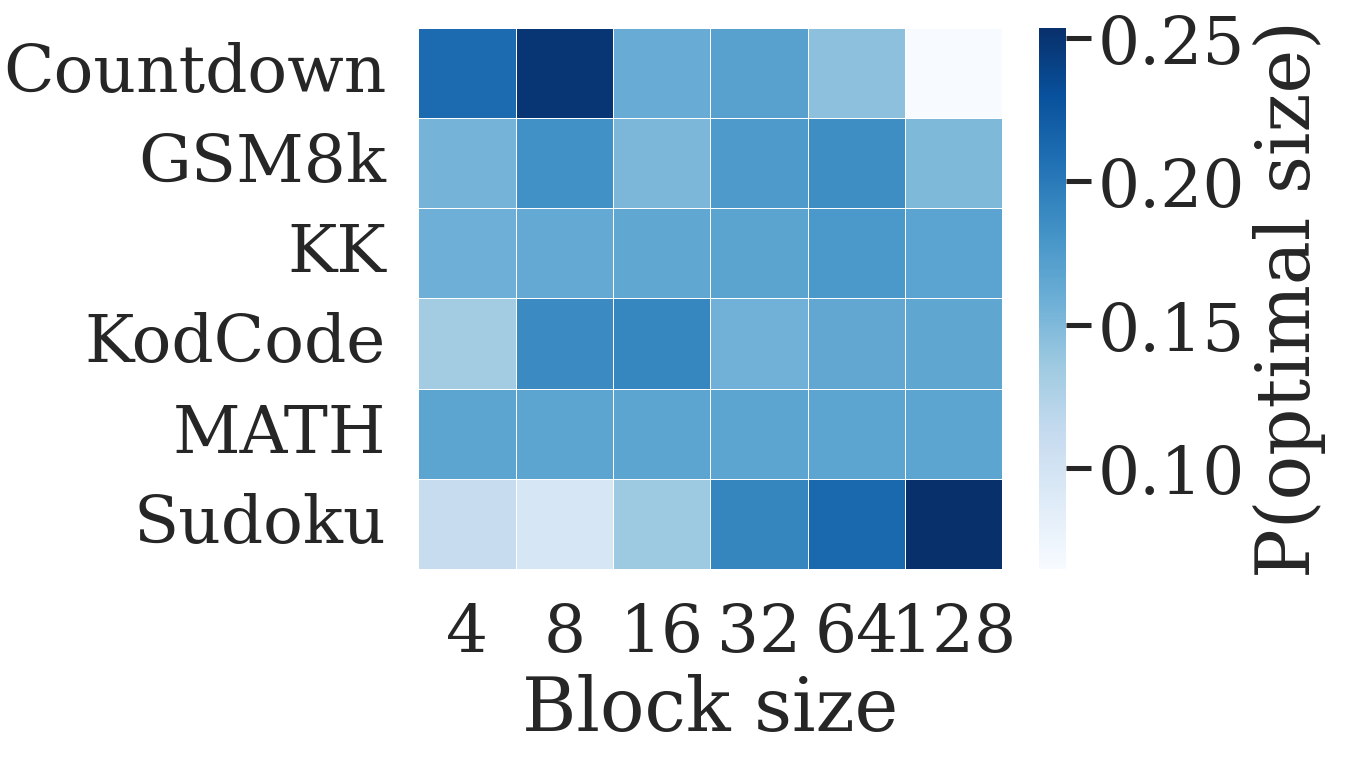}
\vspace{-0.5cm}
\caption{\textbf{Probability distribution of best-improved training block sizes per domain.}
Each cell shows the domain-level training block size preference distribution $P_k^{\mathrm{train}}(c)$ defined in Equation~\ref{eq:domain_dist}. The dLLM is LLaDA2-16B. Darker cells indicate higher probability for the block size to be the best-improved block size.}
\vspace{-0.3cm}
\label{fig:best_block}
\end{wrapfigure}

\subsection{Probability Distribution of Best-Improved Training Block Sizes}
\label{app:best_block}
To further demonstrate domain-level block size preference in dLLM RL post-training, we visualise the probability distribution of domain-level best-improved training block sizes in Figure~\ref{fig:best_block}. Following Equation~\ref{eq:improvement_and_best-improved_c}, each sample $x$ is first assigned a best-improved training block size $c_x^*$ by maximising the teacher-student improvement score over the candidate block size set $B$. Based on this, we compute the empirical probability distribution $P_k^{\mathrm{train}}(c)$ for each domain $\mathcal{D}_k$ following Equation~\ref{eq:domain_dist}. 

As shown in Figure~\ref{fig:best_block}, different domains exhibit clearly different best-improved block size distributions. For example, Countdown assigns higher probability to smaller block sizes such as $4$ and $8$, while Sudoku strongly favours larger block sizes such as $64$ and $128$. This indicates that different reasoning domains prefer different block-based generation structures during dLLM RL post-training. \textbf{Therefore, block size is not a universal fixed hyperparameter, but a domain-dependent structural factor that directly affects effective rollout construction for multi-domain dLLM RL.}

\begin{figure*}[t]
    \centering
    \includegraphics[width=0.9\linewidth]{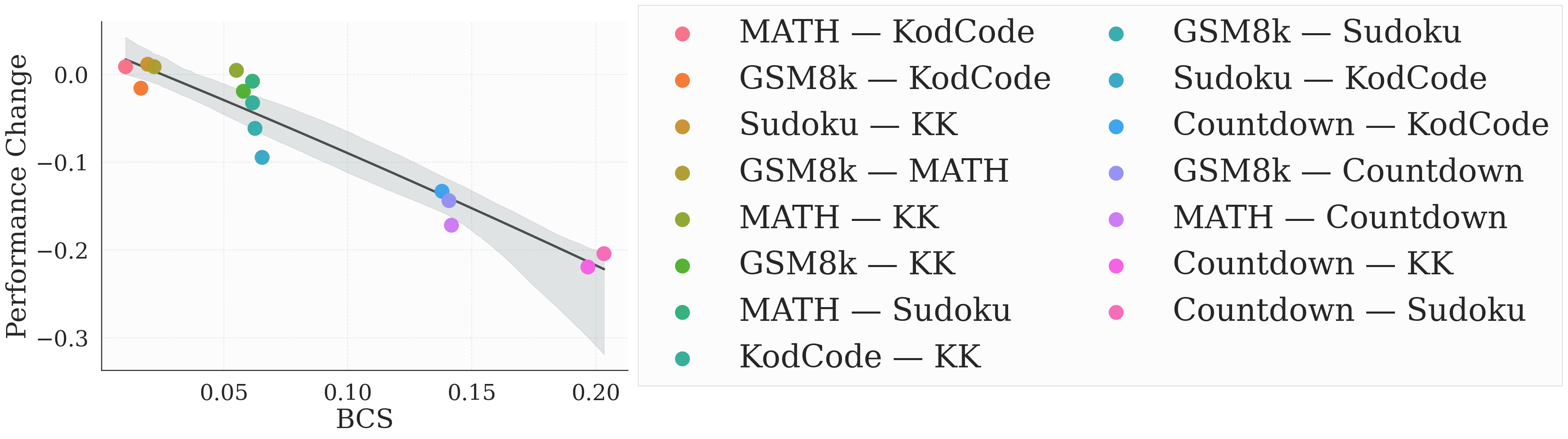}
    \caption{\textbf{Detailed domain-pair legend for BCS analysis.}
    Each point denotes one pair of training domains used for vanilla fixed-block mix-domain RL with StableDRL. The y-axis reports the mean performance change between mix-domain RL and the corresponding single-domain RL results over the two domains.}
    \label{fig:bcs_detail}
\end{figure*}

\newpage
\begin{figure*}[!t]
\centering
\includegraphics[width=1\linewidth]{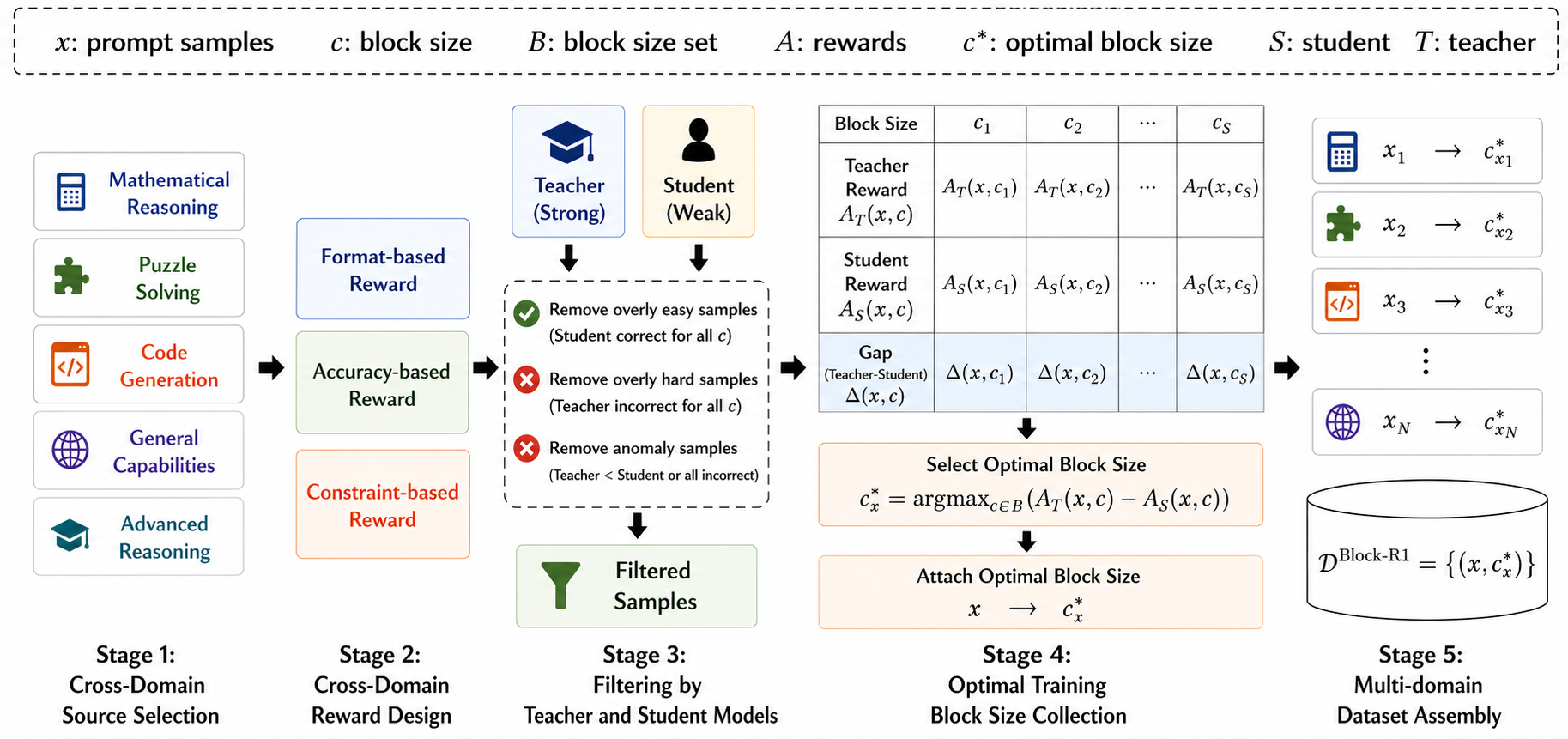}
\vspace{-0.5cm}
\caption{Detailed Illustration of Block-R1-41K Dataset Construction. Block-R1 constructs a block-based multi-domain training corpus through five key stages: cross-domain source selection, unified reward design, teacher-student filtering, sample-level best-improved block size collection, and balanced multi-domain dataset assembly. More details are documented in Algorithm~\ref{app:algorithm}
}
\vspace{-0.5cm}
\label{fig:main}
\end{figure*}
\section{Additional Implementation Details}
\label{app:reproduce}

In this section, we provide additional implementation details. The code is provided at \href{https://github.com/YanJiangJerry/Block-R1}{https://github.com/YanJiangJerry/Block-R1}, with the dataset released at \href{https://huggingface.co/datasets/YanJiangJerry/Block-R1-41K}{https://huggingface.co/datasets/YanJiangJerry/Block-R1-41K}. The detailed dataset construction pipeline is illustrated in Figure~\ref{fig:main}.

\subsection{Benchmark Settings}
\label{app:data}
Block-R1 supports evaluation across diverse benchmark settings spanning code generation, mathematical reasoning, logical puzzles, general capabilities, and advanced reasoning. Specifically, the benchmark includes MBPP~\cite{austin2021program}, HumanEval~\cite{chen2021evaluating}, KodCode~\cite{xu2025kodcode}, GSM8K~\cite{cobbe2021training}, MATH500~\cite{lightman2023let}, Countdown~\cite{tinyzero}, Knights-and-Knaves, Sudoku~\cite{arel_sudoku}, HellaSwag~\cite{zellers2019hellaswag}, MMLU~\cite{hendryckstest2021}, ARC-E and ARC-C~\cite{clark2018think}, and MMLU-Pro~\cite{wang2024mmlupro}. For large datasets, $7{,}000$ training samples are randomly sampled from each domain to ensure balanced data-source sampling, while extremely small domains, such as MBPP with only $374$ training samples, are excluded from the data source for training. KodCode~\cite{xu2025kodcode} is used as the default code generation training set following GDPO~\cite{rojas2025gdpo}. To mitigate potential data redundancy, a consistent deduplication strategy is applied in Block-R1 following GURU~\cite{cheng2025revisiting} by removing the shorter sample if one question is a substring of another.

\begin{table*}[ht]
\centering
\caption{Data Statistics for Block-R1 benchmark. Datasets without an official training split (e.g., HumanEval) are excluded from the source. Note that all domains participate in evaluation. General Capabilities refer to the broad spectrum of knowledge-intensive and commonsense benchmarks that mostly require single-hop logic or fact retrieval. Advanced Reasoning refers to complex multi-step reasoning benchmarks that require deeper knowledge integration and compositional inference.}
\vspace{-0.2cm}
\label{tab:dataset_stats}
\resizebox{0.77\textwidth}{!}{
\begin{tabular}{llcc}
\toprule
\textbf{Category} & \textbf{Dataset} & \textbf{Train Size} & \textbf{Test Size} \\
\midrule
\multicolumn{4}{l}{\textbf{\textit{Code Generation}}} \\
& MBPP & 374 & 500 \\
& HumanEval & N/A & 164 \\
& KodCode & 9,285 & 500 \\
\midrule
\multicolumn{4}{l}{\textbf{\textit{Mathematical Reasoning}}} \\
& GSM8K & 7,473 & 1,319 \\
& MATH500 & 7,500 & 500 \\
& Countdown & 240,632 & 256 \\
\midrule
\multicolumn{4}{l}{\textbf{\textit{Logical Constraint Reasoning (Puzzle Solving)}}} \\
& Knights-and-Knaves (KK) & 6,200 & 700 \\
& Sudoku & 1,000,000 & 256 \\
\midrule
\multicolumn{4}{l}{\textbf{\textit{General Capabilities (Knowledge \& Comprehension)}}} \\
\quad Commonsense Understanding & HellaSwag & 39,905 & 10,003 \\
\quad World Knowledge & MMLU & N/A & 14,042 \\
\quad Scientific Comprehension (Basic) & ARC-E & 2251 & 2,376 \\
\midrule
\multicolumn{4}{l}{\textbf{\textit{Advanced Reasoning (Graduate-Level \& Multi-step Reasoning)}}} \\
\quad World Knowledge (Complex) & MMLU-Pro & N/A & 12,032 \\
\quad Scientific Comprehension (Challenge) & ARC-C & 1119 & 1,172 \\
\bottomrule
\end{tabular}
}
\end{table*}

The dataset statistic is detailed in Table~\ref{tab:dataset_stats}. To ensure a fair comparison, our dataset selection and split follow established protocols from prior dLLM and cross-domain RL studies~\cite{zhao2025d1, tang2025wd1, cheng2025revisiting}. Block-R1 supports benchmark datasets across code generation, mathematical reasoning, logical puzzles, general capabilities, and advanced reasoning. For datasets with official training and test splits, we use their standard partitions whenever available. For datasets without an official training split, the dataset is retained for evaluation.

For mathematical reasoning, GSM8K uses its standard training set and test split, while MATH500 uses the corresponding training partition for post-training and the MATH500 benchmark for evaluation. Countdown uses the training subset from TinyZero~\cite{tinyzero}, with evaluation conducted on 256 synthetic three-number problems following prior work. For logical puzzles, Sudoku uses the $4 \times 4$ dataset with one million training puzzles and 256 synthetic test instances, while Knights-and-Knaves uses its standard training and test partitions. For code generation, KodCode is used as the main training source following GDPO~\cite{rojas2025gdpo}, while MBPP and HumanEval are included for evaluation, with HumanEval treated as evaluation-only because it does not provide an official training split.

For general capability and advanced reasoning evaluation, HellaSwag, ARC-E, and ARC-C follow their standard splits, while MMLU and MMLU-Pro are used as evaluation benchmarks because they do not provide suitable official training splits for our post-training setting. In addition, GURU~\cite{cheng2025revisiting} is further included as a cross-domain RL reference dataset for studying broader cross-domain generalisation. This design ensures that all training data are drawn only from benchmarks with suitable training partitions, while all datasets can still participate in evaluation in Block-R1.

\subsection{Supported dLLM Backbones.}
\label{app:backbone}
Block-R1 supports 10 dLLM backbone models through Hugging Face model ids, including GSAI-ML/LLaDA-8B-Base, GSAI-ML/LLaDA-8B-Instruct, GSAI-ML/LLaDA-1.5, inclusionAI/LLaDA2.0-mini, inclusionAI/LLaDA2.1-mini, Dream-org/Dream-v0-Base-7B, Dream-org/Dream-v0-Instruct-7B, JetLM/SDAR-8B-Chat-b32, Gen-Verse/TraDo-8B-Instruct, and Gen-Verse/TraDo-8B-Thinking. Unless otherwise specified, LLaDA-8B-Instruct~\cite{nie2025large} is used as the default dLLM backbone following most prior dLLM RL methods~\cite{zhao2025d1, tang2025wd1, rojas2025gdpo, wang2026spg, ou2025espo, zhong2026stabilizing}.

\subsection{Baseline Details}
\label{app:baseline}
For RL frameworks, Block-R1 reproduces and supports 7 representative dLLM post-training methods. Token-level GRPO-related methods include \textit{Diffu}-GRPO~\cite{zhao2025d1}, \textit{d1}~\cite{zhao2025d1}, and \textit{wd1}~\cite{tang2025wd1}, where completion-level rewards are applied through token-level likelihood terms or token likelihood reweighting. Sequence-level methods include GDPO~\cite{rojas2025gdpo}, MDPO~\cite{he2025mdpo}, ESPO~\cite{ou2025espo}, and StableDRL~\cite{zhong2026stabilizing}, where the complete denoising trajectory or generated completion is treated as the optimisation unit. Additionally, Block-R1 supports dynamic block size inference by integrating with \textit{b1}~\cite{b1} at inference time. For SFT, we follow the procedure in LLaDA~\cite{nie2025large}, where the model undergoes supervised fine-tuning using the standard denoising cross-entropy objective without RL.

\subsection{Training Protocol}
\label{app:training_protocol}
Our training protocol strictly follows the procedures established in \textit{d1}~\cite{zhao2025d1} and \textit{wd1}~\cite{tang2025wd1}, with GRPO implemented based on the TRL library~\citep{vonwerra2022trl}. For a fair comparison with baselines, all methods are reproduced on four AMD MI300X GPUs, each with 192GB of memory. Low-Rank Adaptation (LoRA) is integrated for GRPO optimisation with rank $r=128$ and scaling factor $\alpha=64$. AdamW~\citep{loshchilov2017decoupled} is used as the optimiser with $\beta_1=0.9$, $\beta_2=0.99$, weight decay $0.1$, learning rate $3\times 10^{-6}$, and gradient clipping threshold $0.2$. A per-device batch size of $12$ and gradient accumulation steps of $1$ are used to maintain a consistent global batch size. To improve computational efficiency, Flash Attention 2~\citep{dao2023flashattention2} and 4-bit quantisation are employed. During gradient update cycles, random masking is applied to each prompt token with probability $p_{\text{mask}}=0.15$ for log-probability estimation. StableDRL~\cite{zhong2026stabilizing} is adopted as the default RL algorithm for all multi-domain experiments in Block-R1 due to its robust empirical performance. More hyperparameter details are in Table~\ref{tab:shared_defaults}. 

\subsection{Evaluation Protocol.}
\label{app:eval_protocol}
We report the best checkpoint performance for each domain following most prior methods~\cite{zhao2025d1, tang2025wd1, rojas2025gdpo}. Specifically, checkpoints are evaluated every $100$ steps within $5{,}000$ global training steps by accuracy (\%) under zero-shot settings. Although some methods~\cite{wang2026spg, zhong2026stabilizing, ou2025espo} originally follow different evaluation settings, such as the 3-shot Sudoku evaluation by SPG~\cite{wang2026spg}, all methods in Block-R1 are evaluated under a consistent zero-shot setting across all domains to ensure a fair comparison. The candidate block size set is set to $\{4,8,16,32,64,128\}$ by default, as these values are commonly used block sizes in prior dLLM methods.

\subsection{Detailed Reward Functions}
\label{app:reward_func}
\textbf{Besides our rewards, the task rewards employed in this study follow those utilised in \textit{d1}~\cite{zhao2025d1} and \textit{wd1}~\cite{tang2025wd1},} with the specific configurations detailed below. Notably, all reward functions, including the reward introduced in \textit{b1}, share the same reward weight 1.
\paragraph{GSM8K.}
We adopt an Unsloth-style composite reward composed of three additive terms:
\begin{itemize}[leftmargin=*, noitemsep]
    \item \textbf{XML Structure Reward} (\texttt{xmlcount\_reward\_func}). The function checks the presence of the four XML markers with exact newline formatting:
    \texttt{<reasoning>\textbackslash n}, \texttt{\textbackslash n</reasoning>\textbackslash n},
    \texttt{\textbackslash n<answer>\textbackslash n}, and \texttt{\textbackslash n</answer>}.
    Each correctly placed marker contributes \(+0.125\), for a maximum of \(0.5\).
    Additionally, small length-based penalties are applied for any extra trailing text after the closing answer tag (a subtraction of \(0.001\) per extra character, with slight variations depending on which split is used).
    \item \textbf{Integer Answer Reward} (\texttt{int\_reward\_func}). After extracting the substring between \texttt{<answer>} and \texttt{</answer>}, the model receives \(+0.5\) if the extracted answer is a valid digit string (i.e., \texttt{.isdigit()}), and \(0\) otherwise.
    \item \textbf{Correctness Reward} (\texttt{correctness\_reward\_func}). The extracted \texttt{<answer>} content is compared against the ground-truth answer string; an exact match yields \(+2.0\), otherwise \(0\).
\end{itemize}
\paragraph{Countdown.}
The reward is computed by extracting an arithmetic expression from the last \texttt{<answer>...\ </answer>} span and verifying both number usage and target correctness:
\begin{itemize}[leftmargin=*, noitemsep]
    \item \textbf{Full Reward} \(+1.0\): if the expression (i) uses \emph{exactly} the provided numbers (multiset match; each number once), (ii) contains only allowed characters \texttt{[\textbackslash d+\textbackslash -*/().\textbackslash s]}, and (iii) evaluates to the target value within tolerance \(10^{-5}\).
    \item \textbf{Partial Reward} \(+0.1\): if an equation is present but either (i) it uses numbers incorrectly, or (ii) it cannot be safely evaluated, or (iii) it evaluates but does not reach the target.
    \item \textbf{Zero Reward} \(0\): if no equation can be extracted from \texttt{<answer>...\ </answer>}.
\end{itemize}
\paragraph{Sudoku.}
The reward measures the fraction of originally empty cells that are filled correctly.
Concretely, the completion is parsed by extracting digits from the last \texttt{<answer>...\ </answer>} span; the resulting digit string is padded/truncated to length 16 for 4\(\times\)4 Sudoku.
Let \(\mathcal{E}\) be the indices of empty cells in the puzzle (cells with \texttt{0}); the reward is
\[
\frac{1}{|\mathcal{E}|}\sum_{i\in \mathcal{E}} \mathbb{1}[\hat{s}_i = s^\star_i],
\]
and is \(0\) if the extracted answer is empty or invalid. This ensures a consistent training setting with most prior RL methods for dLLMs\cite{zhao2025d1, tang2025wd1}.

\paragraph{MATH500.}
Training uses two additive sub-rewards:
\begin{itemize}[leftmargin=*, noitemsep]
    \item \textbf{Format Reward} (\texttt{boxed\_and\_answer\_tags\_format\_reward}). The model is rewarded for producing an \texttt{<answer>...\ </answer>} span and including a \verb|\boxed| expression \emph{inside} that span. Internally, a base score is computed as:
    \begin{enumerate}[label=\roman*), noitemsep]
        \item \(+1.0\) if an \texttt{<answer>} span exists;
        \item plus \(+1.0\) if \verb|\boxed| appears inside that extracted answer span, otherwise \(+0.5\).
    \end{enumerate}
    This base score is then scaled by \(0.5\), yielding the following tiers:
    \begin{enumerate}[label=\roman*), noitemsep]
        \item \(+1.00\) for an \texttt{<answer>} span containing a \verb|\boxed| command;
        \item \(+0.75\) for only an \texttt{<answer>} span;
        \item \(+0.50\) for only the \verb|\boxed| command (without a valid \texttt{<answer>} span);
        \item \(+0.25\) for cases meeting neither condition.
    \end{enumerate}
    \item \textbf{Correctness Reward} (\texttt{correctness\_reward\_func\_math}). The model output is parsed by extracting the \texttt{<answer>} content; the ground-truth string is reduced to the content of its last boxed expression, with \verb|\boxed{...}| removed. The extracted prediction and processed ground truth are then compared using a math-equivalence checker; an equivalent answer yields \(+2.0\), otherwise \(0\).
\end{itemize}
\paragraph{KodCode / MBPP / HumanEval.}
All three code domains use the same two-term reward suite (both with weight 1) during training:
\begin{itemize}[leftmargin=*, noitemsep]
    \item \textbf{Code Format Reward} (\texttt{get\_code\_format\_reward(language="python")}). The completion must contain a Python markdown code block \texttt{```python ... ```}. If no such block exists, reward \(0\). If a code block exists, Python syntax is checked by parsing the extracted code: reward \(+1.0\) if syntactically valid, otherwise \(+0.5\).
    \item \textbf{Execution/Test Reward} (\texttt{code\_reward}). For completions whose format reward is \(1.0\), the extracted code is executed against the provided per-example test cases (\texttt{test\_list}) by running \texttt{python3 -c} for each test case with a per-test timeout of \(30\) seconds. The reward is the pass ratio in \([0,1]\), i.e., \(\text{passed}/\text{total}\). For completions with format reward \(<1\), this term is set to \(0\).
\end{itemize}
\textbf{Note:} while the benchmarks are named MBPP and HumanEval, the training loader in this codebase can reuse a larger code dataset (KodCode-style) while keeping the same reward functions above remain unchanged.
\paragraph{Knights \& Knaves (KK).}
The reward (\texttt{knights\_knaves\_reward\_func}) is a mixture of (i) structured correctness over per-person role assignments and (ii) a small formatting signal:
\begin{itemize}[leftmargin=*, noitemsep]
    \item \textbf{Answer Extraction.} The scorer prefers the text inside \texttt{<answer>...\ </answer>} (with fallbacks such as an \texttt{Answer:} section).
    \item \textbf{Structured Correctness.} It parses lines like ``Alice is a knight/knave'' into a map (\(\text{name}\to\text{role}\)) for both prediction and ground truth. If ground truth assignments are parseable, correctness is the fraction of people whose roles match (a value in \([0,1]\)), otherwise, a normalised-string exact match is used as a fallback.
    \item \textbf{Format Reward (capped).} A small bonus is accumulated for including \texttt{<answer>...\ </answer>}, for having any parseable assignments, for including \texttt{<reasoning>...\ </reasoning>}, and for mentioning keywords ``knight/knave''. This formatting bonus is capped at \(0.3\) to avoid reward hacking issues.
\end{itemize}

\begin{algorithm}[ht]
\caption{\textbf{Block-R1-41K Dataset Construction}}
\label{app:algorithm}
\KwIn{Source domains $\{\mathcal{D}_k\}_{k=1}^{M}$, candidate block sizes $B=\{c_1,\dots,c_S\}$, teacher dLLM $\theta_T$, student dLLM $\theta_S$, reward function $R(\cdot)$, balanced sampling size $n$, rollout number $Q$.}
\KwOut{Block-R1-41K dataset $\mathcal{D}^{\mathrm{Block}\text{-}\mathrm{R1}}$.}

Initialise dataset $\mathcal{D}^{\mathrm{Block}\text{-}\mathrm{R1}} \leftarrow \emptyset$\;

\For{each reasoning domain $\mathcal{D}_k$}{
    Initialise domain-specific pool $\mathcal{X}_k \leftarrow \emptyset$\;
    
    \tcp{\textcolor{blue}{Stage 1: Source Selection and Deduplication}}
    Select samples from $\mathcal{D}_k$ with official training splits\;
    Remove duplicated or highly overlapping samples\;

    \tcp{\textcolor{blue}{Stage 2: Cross-Domain Reward Design}}
    Define domain-specific reward $R(\tau)$ using format-based, accuracy-based, and constraint-based rewards\;

    \For{each sample $x \in \mathcal{D}_k$}{
        Set valid flag $v \leftarrow \mathrm{true}$\;
        
        \tcp{\textcolor{blue}{Stage 3: Teacher-Student Filtering}}
        \For{each block size $c \in B$}{
            Estimate $A_{\theta_T}(x,c)=\mathbb{E}_{\tau \sim \pi_{\theta_T}^{(c)}(\cdot \mid x)}[R(\tau)]$ using $Q$ teacher rollouts\;
            Estimate $A_{\theta_S}(x,c)=\mathbb{E}_{\tau \sim \pi_{\theta_S}^{(c)}(\cdot \mid x)}[R(\tau)]$ using $Q$ student rollouts\;
        }

        \If{student obtains full reward under all $c \in B$}{
            $v \leftarrow \mathrm{false}$\;
        }
        \If{teacher obtains zero reward under all $c \in B$}{
            $v \leftarrow \mathrm{false}$\;
        }
        \If{$\max_{c \in B} A_{\theta_T}(x,c) \leq \max_{c \in B} A_{\theta_S}(x,c)$}{
            $v \leftarrow \mathrm{false}$\;
        }

        \If{$v=\mathrm{true}$}{
            \tcp{\textcolor{blue}{Stage 4: Best-improved Training Block Size Collection}}
            Compute $\Delta(x,c)=A_{\theta_T}(x,c)-A_{\theta_S}(x,c)$ for all $c \in B$\;
            Select $c_x^* \leftarrow \arg\max_{c \in B}\Delta(x,c)$\;
            Add $(x,c_x^*)$ into $\mathcal{X}_k$\;
        }
    }

    \tcp{\textcolor{blue}{Stage 5: Balanced Multi-domain Assembly}}
    Randomly sample $\widetilde{\mathcal{X}}_k \subseteq \mathcal{X}_k$ with $|\widetilde{\mathcal{X}}_k|=\min(n,|\mathcal{X}_k|)$\;
    Add $\widetilde{\mathcal{X}}_k$ into $\mathcal{D}^{\mathrm{Block}\text{-}\mathrm{R1}}$\;
}

\Return{$\mathcal{D}^{\mathrm{Block}\text{-}\mathrm{R1}}$}\;
\end{algorithm}
\subsection{Complexity Analysis}
\label{app:complexity}
The computational overhead of Block-R1 mainly arises from the offline teacher-student evaluation stage during dataset construction. Let $N$ denote the number of candidate training samples after source selection, $S=|B|$ denote the number of candidate block sizes, $Q$ denote the number of rollouts used to estimate each expected reward, $L$ denote the maximum generation length, and $T$ denote the number of diffusion denoising steps. For each sample, Block-R1 evaluates both the teacher and student dLLMs under every candidate block size. Therefore, the total construction cost is $\mathcal{O}(2NSQ \cdot C_{\mathrm{gen}})$, where $C_{\mathrm{gen}}$ denotes the cost of one dLLM generation and reward evaluation. Since dLLM generation is dominated by Transformer self-attention, $C_{\mathrm{gen}}$ can be approximated as $\mathcal{O}(T L^2)$, leading to an overall offline construction cost of $\mathcal{O}(2NSQTL^2)$. In our implementation, $Q=1$ unless otherwise specified, meaning that each $(x,c)$ pair is evaluated with one teacher and student generation. 

The remaining steps introduce only minor overhead. Computing the improvement score $\Delta(x,c)$ and selecting $c_x^*=\arg\max_{c \in B}\Delta(x,c)$ scale as $\mathcal{O}(NS)$ once the teacher-student rewards are obtained, while balanced multi-domain assembly scales linearly with the final dataset size. Importantly, this cost is incurred only once during offline benchmark construction. During RL post-training, Block-R1 only requires an $\mathcal{O}(1)$ lookup of the assigned block size $c_x^*$ for each training sample. Therefore, it does not change the asymptotic complexity of the underlying RL algorithm, while enabling block-conditioned multi-domain rollout generation for dLLMs. This also allows Block-R1 to be freely integrated into various existing RL algorithms for dLLMs without modifying their core optimisation objectives.

\begin{table}[ht]
\centering
\small
\caption{\textbf{Block-R1 adopts a consistent hyperparameter setting to ensure a fair comparison between RL methods.} All experiments use 4$\times$ AMD MI300X GPUs, each with 192GB of memory.}
\label{tab:shared_defaults}
\vskip 0.1in
\begin{tabular}{@{}>{\centering\arraybackslash}p{0.42\linewidth}>{\centering\arraybackslash}p{0.42\linewidth}@{}}
\toprule
\textbf{Parameter} & \textbf{Default Value} \\
\midrule
\multicolumn{2}{c}{\textbf{Model and Precision}} \\
use\_peft & true \\
torch\_dtype & bfloat16 \\
load\_in\_4bit & true \\
attn\_implementation & flash\_attention\_2 \\
lora\_r & 128 \\
lora\_alpha & 64 \\
lora\_dropout & 0.05 \\
peft\_task\_type & CAUSAL\_LM \\
\midrule
\multicolumn{2}{c}{\textbf{Training Configuration}} \\
seed & 42 \\
bf16 & true \\
sync\_ref\_model & true \\
ref\_model\_sync\_steps & 64 \\
adam\_beta1 & 0.9 \\
adam\_beta2 & 0.99 \\
weight\_decay & 0.1 \\
max\_grad\_norm & 0.2 \\
warmup\_ratio & 0.0001 \\
learning\_rate & $3\times 10^{-6}$ \\
lr\_scheduler\_type & constant\_with\_warmup \\
\midrule
\multicolumn{2}{c}{\textbf{Batching / Checkpointing}} \\
per\_device\_train\_batch\_size & 12 \\
per\_device\_eval\_batch\_size & 12 \\
gradient\_accumulation\_steps & 1 \\
dataloader\_drop\_last & true \\
eval\_steps & 100 \\
save\_steps & 500 \\
max\_steps & 5000 \\
log\_completions & false \\
\midrule
\multicolumn{2}{c}{\textbf{RL \& dLLM Hyperparameters}} \\
num\_generations & 6 \\
num\_iterations & 12 \\
max\_completion\_length & 256 \\
max\_prompt\_length & 200 \\
block\_length & 32 \\
diffusion\_steps & 128 \\
generation\_batch\_size & 6 \\
remasking\_strategy & low\_confidence \\
random\_masking & true \\
p\_mask\_prompt & 0.15 \\
beta & 0.04 \\
epsilon & 0.5 \\
\bottomrule
\end{tabular}
\end{table}
\subsection{Reproducibility}
\label{app:code}
The source code is provided to facilitate reproduction with detailed instructions. The main hyperparameter setting follows prior RL for dLLM methods~\cite{zhao2025d1, tang2025wd1, rojas2025gdpo, ou2025espo, zhong2026stabilizing, he2025mdpo} and uses the consistent shared configuration in Table~\ref{tab:shared_defaults}. We further summarise the method-specific hyperparameters below:

\paragraph{WD1.}
WD1 uses its own NSR+PSR objective in the trainer implementation, and the KL coefficient in the shared configuration is not used for KL regularisation in WD1 following its official implementation~\cite{tang2025wd1}.

\paragraph{GDPO and MDPO.}
GDPO and MDPO use a different optimisation schedule to match their original settings. Specifically, the learning rate is set to $3\times 10^{-7}$, the KL coefficient is set to $0.00$, and the maximum training step is increased to $15000$.

\paragraph{ESPO.}
ESPO keeps the shared training defaults but enables sequence-level ELBO estimation through method-specific hyperparameters. The number of Monte Carlo samples is set to $2$, and variance reduction is enabled.

\paragraph{StableDRL.}
StableDRL uses the shared defaults for the main GRPO configuration, while relying on additional StableDRL-specific hyperparameters for SPG and SNIS stabilisation. These settings include enabling SNIS, clipping the importance weight, configuring the SPG coefficients, using one Monte Carlo sample, and adopting block-random forward sampling. These hyperparameters control the SPG/SNIS stabilisation and the SPG bound estimator.

\newpage
\section{Theoretical Analysis}
\label{sec:theoretical_proof}

\begin{tcolorbox}[
  colframe=academicBlue, 
  colback=academicBack,   
  coltitle=white,
  fonttitle=\bfseries, 
  title=Theorem 1: Structural Sub-optimality of Multi-domain RL for dLLMs with Fixed-size Block, 
  boxsep=1pt, left=1pt, right=1pt, top=1pt, bottom=1pt,
  arc=1mm, auto outer arc,
]
Let $c_0 \in B$ denote any globally fixed block size uniformly applied across all domains $\mathcal{D}_k$, and let $\lambda_k>0$ represent normalised sampling weights satisfying $\sum_{k=1}^{M}\lambda_k=1$. Assume $B$ is non-empty and finite so that all maxima over $B$ exist. For any fixed dLLM $\theta$ satisfying Assumptions~1 and 2, the aggregated multi-domain RL objective under the fixed block size $c_0$ is strictly bounded as:
\begin{equation}
\begin{aligned}
\sum_{k=1}^{M} \lambda_k J_k(\theta, c_0) 
< 
\sum_{k=1}^{M} \lambda_k \max_{c \in B} J_k(\theta, c).
\end{aligned}
\end{equation}
\end{tcolorbox}

\begin{proof}
This proof establishes structural sub-optimality over the decoding block-size variable $c$, rather than parameter-level unlearnability over $\theta$. The parameter state $\theta$ is treated as fixed throughout the proof, and the result shows that, under preference divergence, no single shared block size can simultaneously attain the same domain-wise capacity as adaptive block-size selection.

For each domain $\mathcal{D}_k$, let $J_k(\theta,c)$ denote the domain-level reward objective defined in Definition~\ref{def:block_conflict}. Under Assumption~1, this objective is aligned with the expected sample-level block-conditioned reward:
\begin{equation}
\begin{aligned}
J_k(\theta,c)
=
\mathbb{E}_{x \sim \mathcal{D}_k}
\left[
A_\theta(x,c)
\right].
\end{aligned}
\end{equation}

Since $B$ is finite and non-empty, the set of best-improved block sizes for each domain is non-empty. Define the best-improved block-size set of domain $\mathcal{D}_k$ as:
\begin{equation}
\begin{aligned}
C_k^*(\theta)
=
\arg\max_{c \in B} J_k(\theta,c).
\end{aligned}
\end{equation}

By Assumption~2, there exist at least two domains $\mathcal{D}_i$ and $\mathcal{D}_j$ such that their best-improved block-size sets are disjoint:
\begin{equation}
\begin{aligned}
C_i^*(\theta) \cap C_j^*(\theta) = \emptyset.
\end{aligned}
\end{equation}

Now consider any globally fixed block size $c_0 \in B$. Because $C_i^*(\theta)$ and $C_j^*(\theta)$ are disjoint, $c_0$ cannot belong to both sets simultaneously. Therefore, there exists at least one domain $k \in \{i,j\}$ such that:
\begin{equation}
\begin{aligned}
c_0 \notin C_k^*(\theta).
\end{aligned}
\end{equation}

By the definition of the argmax set, if $c_0 \notin C_k^*(\theta)$, then $c_0$ is strictly sub-optimal for domain $\mathcal{D}_k$:
\begin{equation}
\begin{aligned}
J_k(\theta,c_0)
<
\max_{c \in B} J_k(\theta,c).
\end{aligned}
\end{equation}

For every other domain $\mathcal{D}_\ell$, the definition of maximum gives the non-strict inequality:
\begin{equation}
\begin{aligned}
J_\ell(\theta,c_0)
\le
\max_{c \in B} J_\ell(\theta,c).
\end{aligned}
\end{equation}

Multiplying each domain-wise inequality by its positive sampling weight $\lambda_k$ and summing over all $M$ domains yields:
\begin{equation}
\begin{aligned}
\sum_{k=1}^{M}
\lambda_k J_k(\theta,c_0)
<
\sum_{k=1}^{M}
\lambda_k \max_{c \in B} J_k(\theta,c).
\end{aligned}
\end{equation}

The inequality holds because at least one domain has a strict sub-optimal gap and its corresponding weight is positive. This proves that fixed-size block training cannot reach the block-adaptive structural capacity under the stated preference divergence condition. Hence, a one-for-all block size imposes an irreducible structural sub-optimal in multi-domain RL for dLLMs, which completes the proof.
\end{proof}

Therefore, Theorem~\ref{thm:conflict} directly motivates the construction of Block-R1. Since the benchmark assigns sample-specific block sizes, it provides an improvement-aware empirical approximation to the block-adaptive principle in Equation~\ref{eq:theorem_inequality}, rather than forcing all samples to share a single fixed block size. This is precisely why Block-R1 mitigates domain block size conflict in multi-domain dLLM post-training.

\section{Limitations}
\label{app:limitation}
Although Block-R1 provides a systematic benchmark for studying block size adaptation in multi-domain RL for dLLMs, one limitation is that our analysis focuses on block-based dLLMs. Therefore, our method may not directly generalise to other non-autoregressive generation paradigms that do not rely on a block-based decoding structure. Nevertheless, most existing dLLMs adopt or are compatible with block-based inference, where decoding granularity directly affects the generation trajectory. Thus, Block-R1 remains broadly applicable to current dLLM post-training settings while leaving the extension to non-block-based generation frameworks as future work.

\clearpage
\section*{NeurIPS Paper Checklist}

\begin{enumerate}

\item {\bf Claims}
    \item[] Question: Do the main claims made in the abstract and introduction accurately reflect the paper's contributions and scope?
    \item[] Answer: \answerYes{}.
    \item[] Justification: Our paper clearly states the main claims in the abstract and introduction, including the motivation of domain block size conflict, the Block-R1 benchmark construction, the theoretical analysis of fixed-block structural sub-optimality, and the empirical evaluation across multi-domain RL settings for dLLMs.

\item {\bf Limitations}
    \item[] Question: Does the paper discuss the limitations of the work performed by the authors?
    \item[] Answer: \answerYes{}.
    \item[] Justification: Our paper discusses the scope and limitations of our method, including its focus on block-based dLLM post-training, its dependence on the selected candidate block-size set, and the computational cost of teacher-student reward evaluation across multiple domains.

\item {\bf Theory assumptions and proofs}
    \item[] Question: For each theoretical result, does the paper provide the full set of assumptions and a complete (and correct) proof?
    \item[] Answer: \answerYes{}.
    \item[] Justification: Our paper explicitly states the objective alignment and preference divergence assumptions before Theorem~\ref{thm:conflict}. The theorem is stated in Section~\ref{sec:block_r1}, and the complete proof is provided in Appendix~\ref{sec:theoretical_proof}.

\item {\bf Experimental result reproducibility}
    \item[] Question: Does the paper fully disclose all the information needed to reproduce the main experimental results of the paper to the extent that it affects the main claims and/or conclusions of the paper (regardless of whether the code and data are provided or not)?
    \item[] Answer: \answerYes{}.
    \item[] Justification: Our paper provides the key information required to reproduce the main results, including benchmark datasets, source selection, reward design, teacher-student filtering, block-size collection, dLLM backbone, baseline methods, training steps, evaluation protocol, and zero-shot evaluation settings.

\item {\bf Open access to data and code}
    \item[] Question: Does the paper provide open access to the data and code, with sufficient instructions to faithfully reproduce the main experimental results, as described in supplemental material?
    \item[] Answer: \answerYes{}.
    \item[] Justification: Our paper introduces Block-R1 as a benchmark dataset and describes its construction pipeline in detail. We provide access to the benchmark data and code with instructions for reproducing the main experiments in the supplemental material.

\item {\bf Experimental setting/details}
    \item[] Question: Does the paper specify all the training and test details (e.g., data splits, hyperparameters, how they were chosen, type of optimizer) necessary to understand the results?
    \item[] Answer: \answerYes{}.
    \item[] Justification: Our paper specifies the experimental settings, including dataset splits, selected source domains, backbone model, baseline methods, fixed block-size configuration, training steps, checkpoint evaluation interval, batch size, gradient accumulation, generation length, and zero-shot testing protocol.

\item {\bf Experiment statistical significance}
    \item[] Question: Does the paper report error bars suitably and correctly defined or other appropriate information about the statistical significance of the experiments?
    \item[] Answer: \answerYes{}.
    \item[] Justification: Our paper reports standard deviation bars in the corresponding bar plots to show experimental variability where repeated runs are conducted. For the main benchmark tables, following most existing dLLM RL baselines, we report the best checkpoint accuracy under a consistent evaluation protocol. Therefore, standard deviation is not shown for every table, but the reported setting is consistent with prior methods and the available error bars provide evidence of robustness where applicable.

\item {\bf Experiments compute resources}
    \item[] Question: For each experiment, does the paper provide sufficient information on the computer resources (type of compute workers, memory, time of execution) needed to reproduce the experiments?
    \item[] Answer: \answerYes{}.
    \item[] Justification: Our paper reports the main compute configuration used for reproduction, including four AMD MI300X GPUs with 192GB memory each, the per-device batch size, gradient accumulation steps, and global training steps. These details are provided in the experimental setup.
    
\item {\bf Code of ethics}
    \item[] Question: Does the research conducted in the paper conform, in every respect, with the NeurIPS Code of Ethics \url{https://neurips.cc/public/EthicsGuidelines}?
    \item[] Answer: \answerYes{}.
    \item[] Justification: Our research uses existing public benchmark datasets and open dLLM checkpoints for research evaluation. Our paper does not involve human-subject experiments, private user data, surveillance, biometric identification, or other procedures that violate the NeurIPS Code of Ethics.

\item {\bf Broader impacts}
    \item[] Question: Does the paper discuss both potential positive societal impacts and negative societal impacts of the work performed?
    \item[] Answer: \answerYes{}.
    \item[] Justification: Our paper discusses the broader impact of improving dLLM reasoning and multi-domain generalisation. The positive impact is improving efficient and adaptive reasoning systems, while the potential negative impact is that stronger generative reasoning models may be misused if deployed without appropriate safeguards.
    
\item {\bf Safeguards}
    \item[] Question: Does the paper describe safeguards that have been put in place for responsible release of data or models that have a high risk for misuse (e.g., pre-trained language models, image generators, or scraped datasets)?
    \item[] Answer: \answerNA{}.
    \item[] Justification: Our paper introduces a benchmark dataset and evaluation framework rather than releasing a new high-risk pretrained language model or image generator. Block-R1 is constructed from existing public reasoning datasets and does not contain private or sensitive user data.

\item {\bf Licenses for existing assets}
    \item[] Question: Are the creators or original owners of assets (e.g., code, data, models), used in the paper, properly credited and are the license and terms of use explicitly mentioned and properly respected?
    \item[] Answer: \answerYes{}.
    \item[] Justification: Our paper cites the original datasets, dLLM checkpoints, and RL methods used in our experiments. We follow the corresponding usage terms of the existing public assets and document the sources used for constructing and evaluating Block-R1.

\item {\bf New assets}
    \item[] Question: Are new assets introduced in the paper well documented and is the documentation provided alongside the assets?
    \item[] Answer: \answerYes{}.
    \item[] Justification: Our paper introduces Block-R1 as a new benchmark dataset. We document the source selection, reward design, teacher-student filtering, sample-level block-size annotation rule, balanced multi-domain assembly, and evaluation protocol.

\item {\bf Crowdsourcing and research with human subjects}
    \item[] Question: For crowdsourcing experiments and research with human subjects, does the paper include the full text of instructions given to participants and screenshots, if applicable, as well as details about compensation (if any)?
    \item[] Answer: \answerNA{}.
    \item[] Justification: Our paper does not involve crowdsourcing experiments or research with human subjects. All data sources are existing benchmark datasets.

\item {\bf Institutional review board (IRB) approvals or equivalent for research with human subjects}
    \item[] Question: Does the paper describe potential risks incurred by study participants, whether such risks were disclosed to the subjects, and whether Institutional Review Board (IRB) approvals (or an equivalent approval/review based on the requirements of your country or institution) were obtained?
    \item[] Answer: \answerNA{}.
    \item[] Justification: Our work does not involve human-subject research, user studies, crowdsourcing, or collection of personal information. Therefore, IRB approval is not applicable.

\item {\bf Declaration of LLM usage}
    \item[] Question: Does the paper describe the usage of LLMs if it is an important, original, or non-standard component of the core methods in this research? Note that if the LLM is used only for writing, editing, or formatting purposes and does \emph{not} impact the core methodology, scientific rigor, or originality of the research, declaration is not required.
    \item[] Answer: \answerYes{}.
    \item[] Justification: Our paper explicitly studies dLLMs and describes the use of frozen teacher and student dLLM checkpoints in the Block-R1 construction pipeline. The dLLM backbone, teacher-student evaluation process, and dLLM-based RL algorithms are central components of our method and are described in the main paper.

\end{enumerate}

\end{document}